\documentclass[10pt,twocolumn,letterpaper]{article}

\usepackage{iccv}              

\definecolor{iccvblue}{rgb}{0.21,0.49,0.74}
\usepackage[pagebackref,breaklinks,colorlinks,allcolors=iccvblue]{hyperref}

\usepackage{multirow, mathrsfs, booktabs,color, amsfonts, amsmath, bbding,bm}
\usepackage{algorithm}
\usepackage{algorithmic}
\usepackage{newfloat}
\usepackage{xcolor}
\usepackage{listings}

\def\paperID{5348} 
\def\confName{ICCV}
\def\confYear{2025}

\title{Learnable Fractional Reaction-Diffusion Dynamics for \\ Under-Display ToF Imaging and Beyond}

\author{Xin Qiao$^1$ \hspace{0.7cm} Matteo Poggi$^2$ \hspace{0.7cm} Xing Wei$^3$ \\ Pengchao Deng$^1$  \hspace{0.7cm} Yanhui Zhou$^{1,}\thanks{Corresponding author}$\hspace{0.7cm} Stefano Mattoccia$^2$ \\
$^1$Xi'an Jiaotong University \hspace{0.7cm} $^2$University of Bologna \hspace{0.7cm} $^3$Anyang Institute of Technology 
}

\begin{document}
\maketitle
\begin{abstract}

Under-display ToF imaging aims to achieve accurate depth sensing through a ToF camera placed beneath a screen panel. However, transparent OLED (TOLED) layers introduce severe degradations—such as signal attenuation, multi-path interference (MPI), and temporal noise—that significantly compromise depth quality.
To alleviate this drawback, we propose \textbf{L}earnable \textbf{F}ractional \textbf{R}eaction-\textbf{D}iffusion \textbf{D}ynamics (LFRD$^2$), a hybrid framework that combines the expressive power of neural networks with the interpretability of physical modeling. 
Specifically, we implement a time-fractional reaction-diffusion module that enables iterative depth refinement with dynamically generated differential orders, capturing long-term dependencies. In addition, we introduce an efficient continuous convolution operator via coefficient prediction and repeated differentiation to further improve restoration quality. Experiments on four benchmark datasets demonstrate the effectiveness of our approach. 
The code is publicly available at \href{https://github.com/wudiqx106/LFRD2}{https://github.com/wudiqx106/LFRD2}.

\end{abstract}

\section{Introduction}

The ascendancy of full-screen featuring a high screen-to-body ratio within the realm of intelligent terminals (e.g. smartphones) design has marked a significant leap forward in enhancing both user experience and aesthetic appeal.
This trend underscores the importance of integrating cameras to facilitate an immersive and visually captivating interactive environment.
Recently, extensive research \cite{zhou2021image,feng2021removing} on under-display RGB image restoration has been conducted by both academia and industry, leading to its successful deployment in the front cameras of mass-produced smartphones.
Furthermore, as the pursuit of a truly seamless display experience advances, under-display depth cameras, such as under-display ToF (UD-ToF) sensors \cite{qiao2022depth}, have drawn considerable interest.
This technology, capable of capturing three-dimensional spatial information through the screen panel, represents a crucial step forward in the evolution of intelligent terminal design.
Indeed, compared to under-display RGB imaging, Transparent Organic Light-Emitting Diode (TOLED) panels pose more severe challenges for ToF cameras, such as reduced ranging accuracy and the loss of depth details, among others. 

\begin{figure}[t]          
	\centering
	\setlength{\abovecaptionskip}{0pt}	
	\captionsetup[subfigure]{font={scriptsize,stretch=1.25},justification=centering}	
	\subfloat[]{
		\centering
		\includegraphics[height=0.93in]{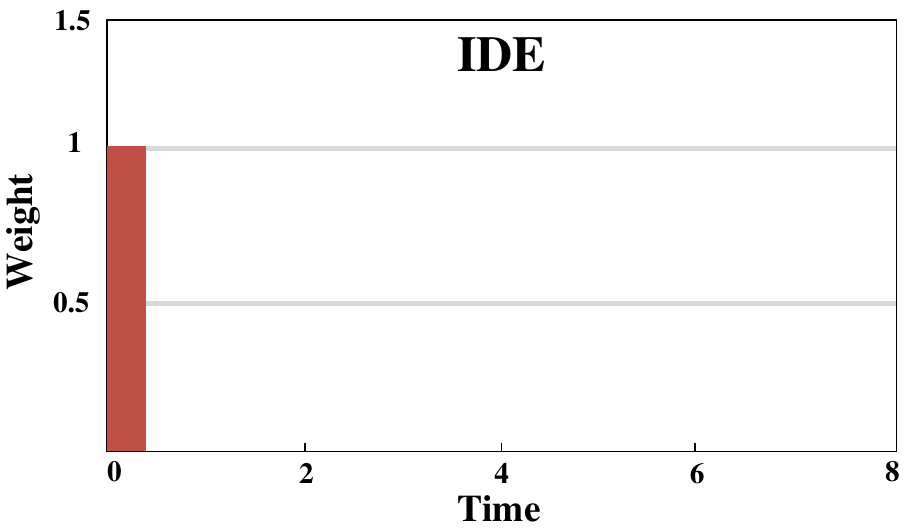}
		\label{ide}
	}
	\subfloat[]{
		\centering
		\includegraphics[height=0.93in]{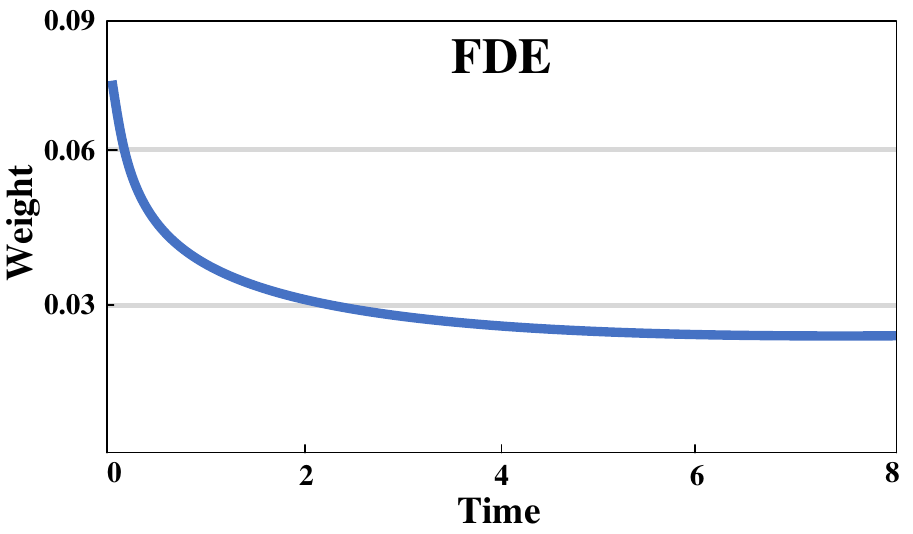}
		\label{fde}
	}
        
        \vspace{0mm}
        \subfloat[]{
		\centering
		\includegraphics[height=0.90in]{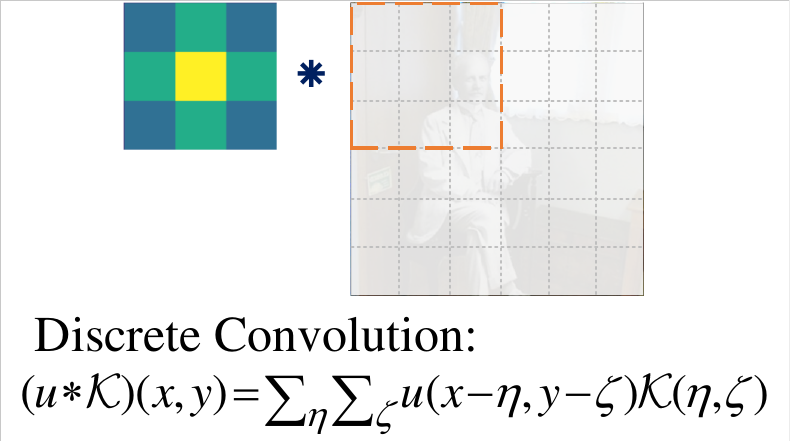}
		\label{dis_conv}
	}
	\subfloat[]{
		\centering
		\includegraphics[height=0.90in]{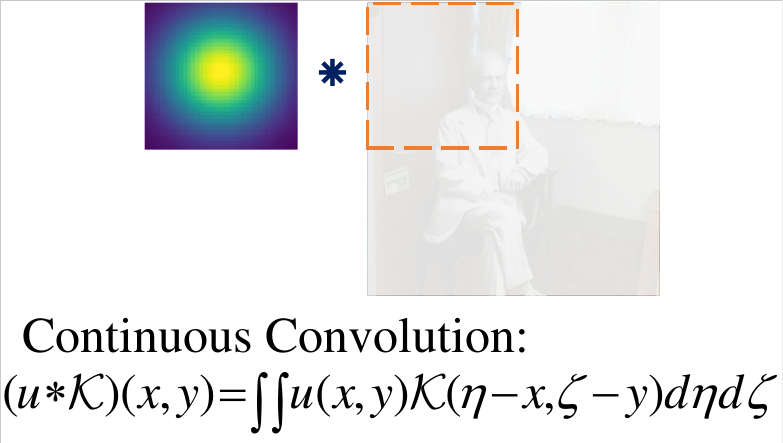}
		\label{conts_conv}
	}

        \vspace{0mm}
	\subfloat[]{
		\centering
		\includegraphics[height=0.97in]{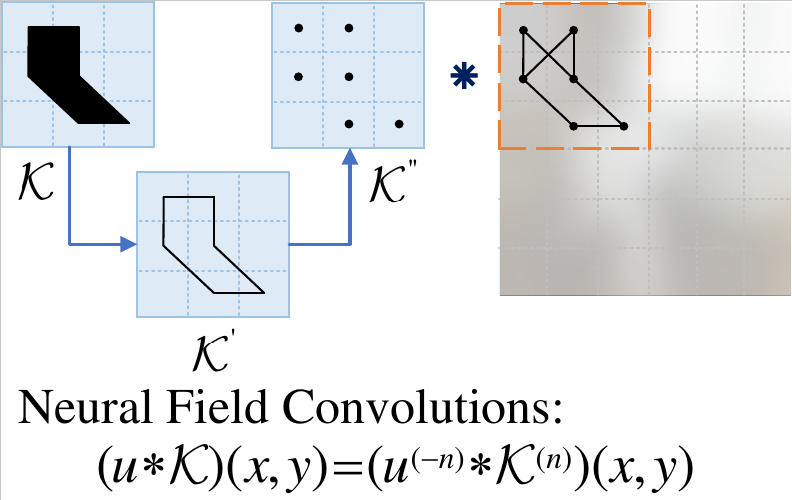}
		\label{NFC}
	}
	\subfloat[]{
		\centering
		\includegraphics[height=0.97in]{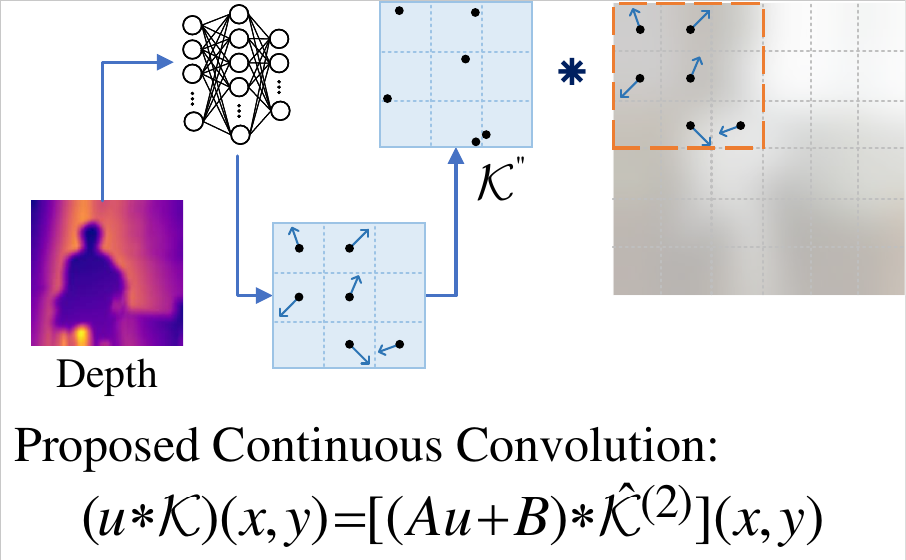}
		\label{our_cc}
	}	
	\caption{\textbf{Core components in LFRD$^2$.} On top: comparison between (a) Integer Differential Equation (IDE), (b) Fractional Differential Equation (FDE); At bottom: comparison between (c) Discrete Convolution and (d) Continuous Convolution, followed by some implementations of the latter, i.e., (e) Neural Field Convolution, and (f) ours.}
	\label{teaser}
\end{figure}

In addressing these issues, classical diffusion processes that leverage domain-specific physical priors, such as Perona-Malik (P-M) diffusion \cite{perona1990scale} and its variants~\cite{scharstein1998stereo}, are recognized as robust tools for improving depth accuracy and preserving details.
Mathematically, given the spatial information $\mathbf{z}$ on a bounded domain $\Omega \in \mathbb{R}^2$ and time $t$, the latent clear image $u(\mathbf{z})$ can be obtained by solving the equation
\begin{equation}
    \label{pm_model}
    \left\{
    \begin{aligned}
        &\frac{\partial u}{\partial t}=\mathrm{div}(g|\nabla  u|\nabla  u), \quad &&(\mathbf{z}, t) \in \Omega \times (0, T] \\
        &u(\mathbf{z}, 0)=u_0(\mathbf{z}),  &&\mathbf{z} \in \Omega \\
    \end{aligned}
    \right.
\end{equation}
with $u_0(\mathbf{z})$ being the initial condition, $g(\cdot)$ the diffusivity function and $u=u(\mathbf{z}, t)$ the solution at time $t$. 
These methods exhibit robust adaptability and generalization, yet the requirement for modeling numerous and complex parameters, extensive computational demands, and the neglect or misassumptions of non-primary factors in imaging degradation burden them.
In this context, deep learning methods, with their strengths in high-level image understanding and contextual reasoning, have garnered considerable attention as potential solutions.
However, their reliance on the meticulous design of network architectures, as well as the abundance and quality of data, remains a significant factor.

Notably, efforts~\cite{cheng2019learning,metzger2023guided} to establish specific and systematic constraints between commonly used iterative algorithms in diffusion processes and deep neural networks -- also known as algorithm unrolling -- have shown promising advances in depth restoration.
Nonetheless, these iterative methods typically employ integer differential equations (IDE), where the predicted state $u_{n+1}$ depends solely on the current state $u_n$, as illustrated in Fig.~\ref{ide}.
In real systems, the predicted state often depends not only on the current state but also on previous ones.
Fig.~\ref{fde} further illustrates the memory properties inherent to fractional-order dynamic systems, which capture this historical dependency.
Some traditional methods~\cite{guo2021three,lian2023non} have leveraged this property to advance image processing, whereas challenges in analytical solutions and parameter selection persist in fractional-order systems.
This difficulty motivates us to harness the fitting capabilities of neural networks to approximate the solutions, thereby improving UD-ToF imaging quality.

In a diffusion step for image processing, the central pixel is updated based on a weighted combination of its neighboring pixels, effectively performing a discrete convolution to propagate information.
However, scenes in the natural world are continuous rather than discrete, sparking significant interest in neural fields \cite{xie2022neural,shue2023d,vasconcelos2023cuf}.
Also known as implicit neural representations, neural fields are typically implemented by Multi-Layer Perceptrons (MLPs) which learn a continuous function mapping spatial coordinates into signals or kernels \cite{shue2023d,vasconcelos2023cuf}, 
these latter to replace the widespread discrete convolutions with continuous ones -- both illustrated in Fig.~\ref{dis_conv} and \ref{conts_conv}. 
Despite demonstrating promising prospects in tasks such as 3D reconstruction and image super-resolution, they are still hindered by drawbacks like high computational costs and intricate hyperparameters tuning. 
Differently, using repeated differentiation~\cite{nsampi2023neural}, shown in Fig.~\ref{NFC}, offers a pragmatic strategy for efficiently implementing continuous convolution, but its flexibility is limited by placing control points on fixed grids and predefining the convolution kernel.

In this paper, we develop a hybrid approach termed \textbf{L}earnable \textbf{F}ractional \textbf{R}eaction-\textbf{D}iffusion \textbf{D}ynamics (LFRD$^2$), which combines neural networks with physical modeling in an end-to-end training framework, enabling depth optimization in a coarse-to-fine manner. 
Here, the neural networks embedded within the time-fractional reaction-diffusion equation learn to optimize the iterative errors generated at each step based on previous states, rather than functioning as an end-to-end regressor.
Notably, the differentiation orders are no longer fixed at predetermined values but are dynamically generated by a neural network.
During the diffusion process, we propose a novel method for continuous convolution based on the properties of signal convolution, illustrated in Fig.~\ref{our_cc}, which can efficiently improve depth quality.
This approach, simply leveraging several flat convolution layers instead of coordinate-based MLPs, achieves continuous convolution via parameter prediction while offering robust interpretability.
This design enables efficient, interpretable continuous convolution, and offers potential extensibility to other depth-related tasks.
Fig.~\ref{teaser} highlights the main properties differentiating our proposal concerning existing methods. Accordingly, our main contributions can be summarized as follows:
\begin{itemize}
    \item {We present a hybrid framework that integrates neural networks into a learnable fractional reaction-diffusion equation, leveraging prior physics knowledge to iteratively refine depth and enable effective learning with variable fractional order.}
    \item {We introduce an efficient continuous convolution operator that leverages coefficient prediction and repeated differentiation, boasting robust interpretability alongside parameter efficiency.}
    \item {The proposed framework was evaluated on two UD-ToF and two depth restoration benchmark datasets, confirming its theoretical and experimental consistency, and validating its effectiveness in UD-ToF imaging and beyond.}
\end{itemize}

\section{Related Work}

We briefly review the literature relevant to our proposal.

\textbf{Under-Display Sensor Imaging.}
Existing under-display sensor imaging is mainly divided into two types: Under-Display RGB and Under-Display ToF. Among them, the former was developed earlier. The optical system of an under-display RGB camera is analyzed for the first time by \citet{zhou2021image}, who also present a dataset for this analysis. The Point Spreading Function (PSF) of the under-display device is directly measured by utilizing a point light source \cite{feng2021removing}, with this measurement being integrated as a pivotal component within their data synthesis process. A novel degradation model for under-display imaging is proposed by \citet{koh2022bnudc}, taking into account the color shift and signal attenuation that vary across different positions on the Transparent OLED screen. 
To address the challenge of low-contrast image enhancement, statistical properties of the H and S channels in HSV space of under-display images are analyzed, and a pixel-level estimation network is proposed by \citet{luo2022under}.
\citet{feng2023generating} design an innovative Transformer-based architecture to alleviate the non-negligible domain discrepancy and spatial misalignment, resulting in superior-quality target data.
Recently, \citet{liu2023fsi} proposed a network architecture that incorporates interactive learning between frequency and spatial domains to mitigate the effects of various scattering phenomena.
\citet{li2023lightweight} design a lightweight network to estimate distortion-free images by leveraging wavelet transformation and multi-scale feature fusion. Since these methods fail to consider the physical correlations in ToF raw data, they cannot be directly applied to UD-ToF imaging.

A few frameworks were also designed to solve depth restoration for under-display ToF. 
The pioneering work~\cite{qiao2022depth} focuses on a depth restoration framework and synthetic data algorithm, tailored to overcome complex degradation in ToF imaging through Transparent OLED displays.
A similar work~\cite{sun2023under} utilizes an optimized Restormer~\cite{zamir2022restormer} to replace the second stage lightweight network by \citet{qiao2022depth}.
Although these methods have achieved remarkable progress in under-display ToF depth restoration, they do not involve research on interpretability.

\textbf{Nonlinear Diffusion for Image Enhancement.}
Nonlinear diffusion has been widely applied to address image enhancement. Traditional works usually leverage mathematical models to generate result images. 
To remove the noise of the image, a new model \cite{liao2021time} based on the time-fractional diffusion equation is proposed, which is stable and the numerical solution converges.
A denoising model \cite{li2023fractional} is built based on fractional-order and integer-order diffusions, taking advantage of texture-preserving and edge-preserving properties.
For image denoising and restoration, a novel partial differential equation, utilizing a time-fractional order derivative, is proposed by \citet{ben2023regularized}.
Recently, a coupled nonlinear diffusion system \cite{du2024nonlinear} is designed to both restore and binarize a degraded document image.

Since deep learning has become widely popular in various fields of image processing, researchers concentrate on incorporating diffusion models with deep learning methods. 
The trainable dynamic nonlinear reaction-diffusion model, featuring time-dependent filter parameters and influence functions learned from data, is introduced by \citet{chen2016trainable}.
A denoising network \cite{jia2019focnet} based on the discretization of a fractional-order differential equation is developed to consider long-term memory in both forward and backward passes.
Moreover, \citet{metzger2023guided} leverage a novel approach utilizing guided anisotropic diffusion with a deep convolutional network for guided depth super-resolution.
However, these methods either fail to account for the influence of previous states during the iteration, or expose unclear statistical or physical explainability.

\textbf{Continuous Convolution.}
Continuous convolution has become popular in computer vision due to its advantages in handling irregular data and preserving original information.
In 3D vision, increasing approaches adopt the continuous kernel to improve the quality of irregular 3D point clouds~\cite{wang2018deep,mao2019interpolated,thomas2019kpconv}.
A small neural network can represent a convolutional kernel as a continuous function, enabling the parallel processing of arbitrarily long sequences within a single operation \cite{romero2021ckconv}
A new approach \cite{kim2023smpconv} dynamically adjusts parameters, enabling continuous function construction via interpolation, which achieves a lightweight structure with enhanced performance.
Recently, 
\citet{nsampi2023neural} propose to train a repeated integral field, requiring only a small number of point samples from the neural integral field to perform an exact continuous convolution. 
However, these methods bring high computational costs and intricate hyperparameter tuning.

\section{Methodology}
\begin{figure*}[ht]
\begin{center}
\includegraphics[width=6in]{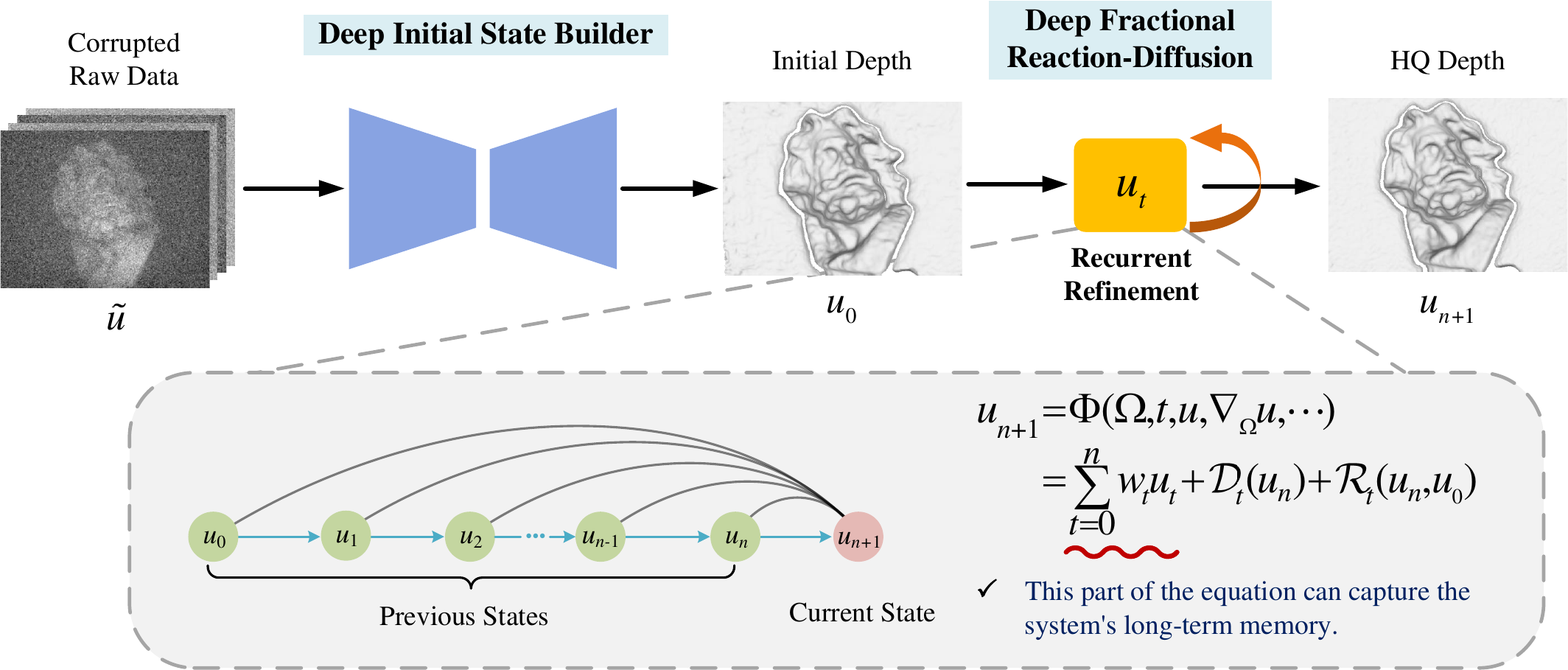}
\end{center}\vspace{-0.5cm} 
  \caption{\textbf{Overview of LFRD$^2$.} Our framework deploys a Deep Initial State Builder, which can be any among the existing networks for UD-ToF imaging, to obtain an initial depth map. Then, the Deep Fractional Reaction-Diffusion module iteratively optimizes it to obtain the final, high-quality (HQ) depth map. }\vspace{-0.2cm} 
\label{fig:framwork}
\end{figure*}

UD-ToF imaging strives to yield high-quality depth maps from corrupted raw measurements. 
To achieve this, we introduce our learnable fractional reaction-diffusion framework, comprised of two novel components. 
First, the overall paradigm of the proposed framework in Fig. \ref{fig:framwork} is presented.
Then we elaborate on the fractional reaction-diffusion dynamics with the derivation of underlying physics.
Finally, the efficient continuous convolution operator is illustrated.

\subsection{Proposed Framework}

As shown in Fig.~\ref{fig:framwork}, our framework primarily encompasses two processes: deep initial state builder (DISB) and deep fractional reaction-diffusion. 
DISB is introduced to generate the initial state $u_0$ for subsequent iterative depth refinement.
For denoising, it serves as a denoising network or an identity mapping, while for image super-resolution, it functions as an upsampling network.
In the deep fractional reaction-diffusion process, we combine the time-fractional reaction-diffusion equation with neural networks to iteratively refine the depth map, where $u_t$ denotes the intermediate depth at iteration $t$.
Since each current state depends on all previous ones, the recurrent refinement captures long-term memory. Mathematically, the evolution of this process is expressed as:
\begin{equation}
\label{eq_fdiffusion}
    \begin{aligned}
        u_{n+1}&=\Phi(\mathbf{z},t,u,\nabla_{\mathbf{z}}u, \cdots)  \\
          &=\sum_{t=0}^n w_tu_t+\mathcal{D}_{t}(u_n)+\mathcal{R}_{t}(u_n, u_0)
    \end{aligned}
\end{equation}
where $\Phi$ represents a nonlinear operator that characterizes the right-hand side of the PDE, $w_t$ indicates the memory weight of the previous state in defining the present stage at time $t$, and $\nabla_{\mathbf{z}}$ is the gradient operator in spatial information. The $\mathcal{D}_{t}(\cdot)$ and $\mathcal{R}_{t}(\cdot)$ denote the diffusion term and reaction term, respectively. The detailed description of Eq.~(\ref{eq_fdiffusion}) will be provided in the subsequent section.

\subsection{Fractional Reaction-Diffusion Dynamics}
IDEs, such as the P-M model, rely solely on the current state for prediction during iteration, often leading to blur and artifacts~\cite{lian2023non} -- for further details, see the \textbf{supplementary material}. In contrast, FDEs benefit from long-term dependence, accumulating historical information over iterations to better mitigate these drawbacks.
Furthermore, the non-local properties of fractional FDEs offer a suitable framework for explaining dynamic processes that exhibit memory effects, thereby enhancing the description of real-world physical phenomena. 
In practice, three commonly used fractional-order derivatives are the Riemann-Liouville, Caputo, and Grünwald-Letnikov formulations, each exhibiting distinct characteristics in numerical computations. 
Among them, the Caputo derivative stands out for its clear physical interpretation and straightforward initial condition handling, making it suitable for physical and engineering modeling. Therefore, we adopt the Caputo derivative.
For an order $\alpha$ ($0 < \alpha < 1$) and a state $u(t)$, the Caputo derivative $D_t^{\alpha}u(t)$ can be expressed as:
\begin{equation}
    ^{C}_0D_t^{\alpha} u(t) = \frac{1}{\Gamma(1-\alpha)} \int_0^t (t-\tau)^{-\alpha} u^{\prime}(\tau) \, d\tau
\end{equation}
where $\Gamma(\cdot)$ represents the Gamma function, while $u^{\prime}(t)$ denotes the first-order derivative of the state $u(t)$. The integral is evaluated over the interval from 0 to $t$.

To solve fractional-order differentials, we discretize the Caputo derivative with $L1$ approximation. The formula at $t=t_{n+1}$ can be approximated as:
\begin{equation}
\label{eq_pfe}
   ^C_0  D_t^{\alpha}u_{n+1} \approx \frac{(\Delta t)^{-\alpha}}{\Gamma(2-\alpha)} \sum_{k=0}^{n}a_k^{(\alpha)}[u_{n+1-k}-u_{n-k}]
\end{equation}
where $u_i=u(t_i,\mathbf{z})$, $i=0,1,2,...,n$, $a_k^{(\alpha)}=(k+1)^{1-\alpha}-k^{1-\alpha}, l\geq 0$. $(\cdot)^{(\alpha)}$ represents the derivative, distinguishing it from the power index.  

Here, we choose the $L$1 approximation over other ones for two main reasons \cite{mustapha2020l1}: firstly, the estimation accuracy from the $L1$ approximation is adequate for our purposes, and secondly, the $L$1 approximation involves a lower computational complexity.
From the physical perspective, the nonlinear fractional reaction-diffusion process can be formulated in an explicit numerical scheme as:
\begin{equation}
    \label{eq_state}
    ^{C}_{0}D_{t}^{\alpha}u_{n+1}=\text{div}(g(|\nabla u_n|)\nabla u_n)+\lambda (u_0-u_n)
\end{equation}
where $\text{div}(g(|\nabla u_n|)\nabla u_n$ is the Perona-Malik diffusion process $\mathcal{D}_{t}(\cdot)$ \cite{perona1990scale}, and $\lambda (u_0-u_n)$ is the reaction term $\mathcal{R}_{t}(\cdot)$, also known as the additional bias term, which serves to drive the depth evolution toward a target state while preserving essential features of the source. 
Here, $g(\cdot)$ is generated by a neural network with flat convolution layers, rather than derived from a conductance function~\cite{perona1990scale}.
Following TNRD~\cite{chen2016trainable}, we set $\lambda=0.01$.

When $\Delta t=1$, we can derive the physical model-driven refinement by combining Eq. (\ref{eq_pfe}) and Eq. (\ref{eq_state}):
\begin{equation}
    \begin{aligned}
    u_{n+1}&=u_n+ S \left[\text{div}(g|\nabla u_n|\nabla u_n)+\lambda (u_0-u_n))\right] \\
     &-\left[\sum_{k=1}^{n} a_k^{(\alpha)}(u_{n+1-k}-u_{n-k})\right]
    \end{aligned}
\end{equation}
with $S=\frac{\Gamma (2-\alpha)}{a_0^{\alpha}}$.
Building on the work~\cite{ashurov2020determination} of Ashurov et al., the fractional order in the subdiffusion equation is guaranteed to exist and be unique when the initial condition is given (i.e., the depth map output by DISB) and appropriate boundary conditions (i.e., Neumann boundary condition~\cite{rao2023encoding}) and constraints are imposed. While noise may affect the numerical stability of the solution, it does not compromise its existence or uniqueness. This suggests that neural networks, as a flexible and efficient alternative to traditional numerical methods, can be leveraged to estimate the fractional order, which can be viewed as a form of physics-informed neural networks (PINNs).

The iterative refinement integrates the intrinsic physical information within the learnable diffusion dynamics \cite{rao2023encoding}, thereby endowing the entire process with interpretability and aligning it closely with the physical process.

\begin{figure}[t]

\begin{center}
\includegraphics[width=3.3in]{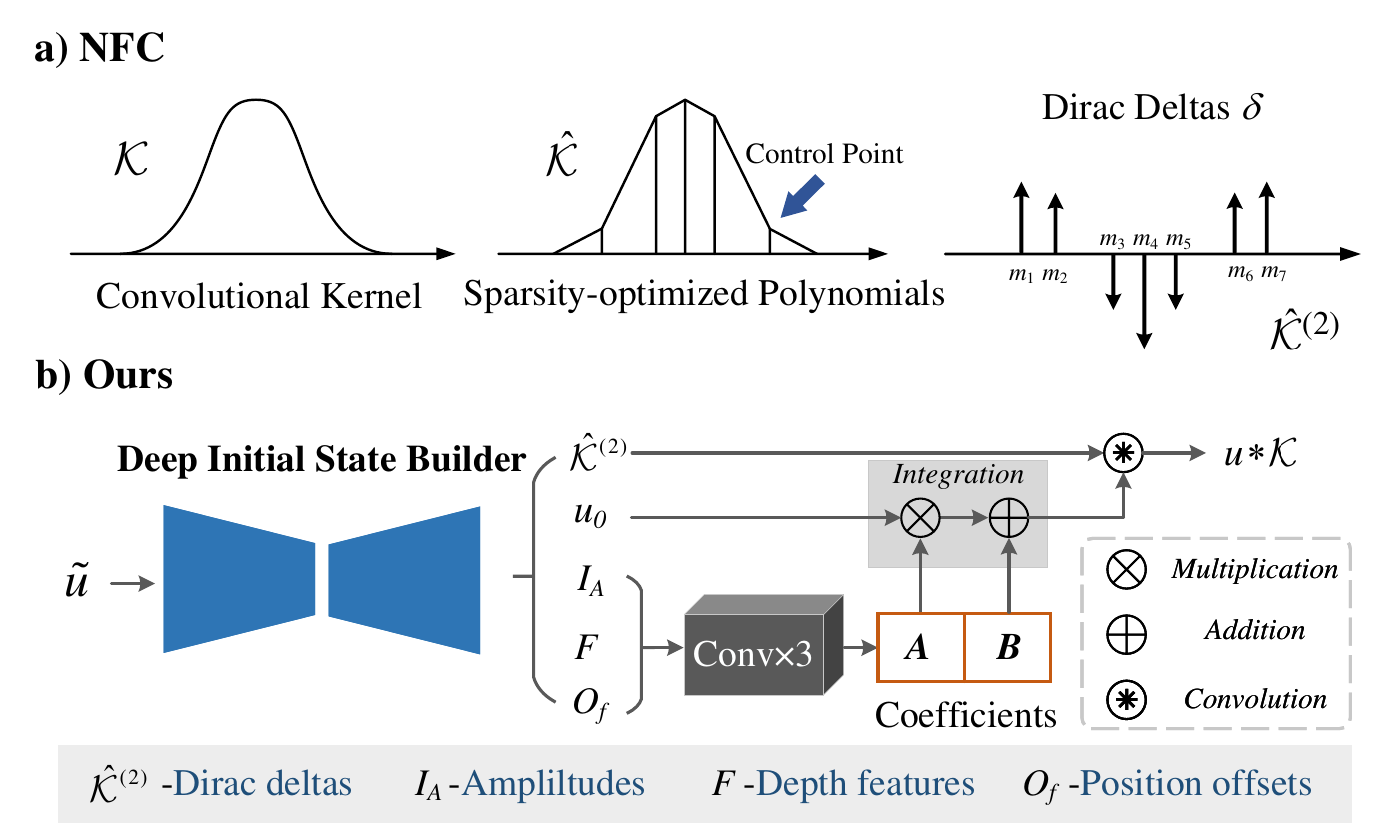}
\end{center}
  \caption{\textbf{Overview of our Continuous Convolution module.} Comparison between NFC and our proposal.}
\label{fig:CC2d}
\end{figure} 

\subsection{Continuous Convolution}

In iterative optimization, each step of image diffusion can be interpreted as a discrete convolution, which overlooks the inherent continuity of natural scenes.
To address this, we implement $div(g(|\nabla u|)\nabla u)$ using a learnable continuous convolution operator, enabling flexible and data-adaptive spatial propagation.
This motivation aligns with recent advances in neural fields, where continuous convolution is often implemented via MLPs to approximate coordinate-based functions, though such designs suffer from high computational cost and complex tuning.
Employing repeated differentiation and integration~\cite{heckbert1986filtering}, as shown in Fig.~\ref{fig:CC2d}$(a)$, can be an efficient pathway towards achieving continuous convolution~\cite{nsampi2023neural}:
\begin{equation}		
		\resizebox{0.9\hsize}{!}{$\begin{aligned}
		\label{eq_convolution}
            u*\mathcal{K}=\underbrace{\left(\int^{n}\ldots \int^{n}u\text{d}\mathbf{z}_1^{n}\ldots \text{d}\mathbf{z}_d^{n}\right)}_{u^{(-n)}}*\underbrace{\left(\frac{\partial^{dn}}{\partial \mathbf{z}_1^{n}\ldots\partial \mathbf{z}_d^{n}}\mathcal{K} \right)}_{\mathcal{K}^{(n)}}
		\end{aligned}$}
\end{equation}
where $u$ and $\mathcal{K}$ represent signals and kernels, while $(\cdot)^{(-n)}$ and $(\cdot)^{(n)}$ denote, respectively, multidimensional repeated antiderivatives and derivatives, and $n$ is the number of repeated operations.

\begin{table*}[t] \small     
	\renewcommand\tabcolsep{4pt} 
	\centering	
	\begin{tabular}{@{}cccccccccccc@{}}
		\toprule
		\multicolumn{1}{c}{\textbf{Dataset}} & \multicolumn{1}{c}{\textbf{Metrics}} & CDNLM & JGDR & ToFnet & ToF-KPN & SHARPnet & PE-ToF & NAFNet & Restomer & UD-ToFnet & LFRD$^2$ \\ \midrule
		& Input & Raw & Depth & Raw & Depth & Depth & Raw & Depth & Depth & Raw & Raw \\ \midrule
		\multirow{5}{*}{\rotatebox[origin=l]{90}{\textbf{SUD-ToF}}}
		& MAE$\downarrow$ & 33.23 & 9.14 & 10.34 & 13.39 & 14.84 &  9.77 & 11.08 & 9.75 & \underline{8.88} & \textbf{8.41} \\
		& RMSE$\downarrow$ & 48.43 & 34.43 & 28.28 & 21.05 & 23.00 &  15.92 & 18.24 & 14.76 & \underline{11.50} & \textbf{10.99}  \\
		& $\rho_{1.02}\uparrow$ & 51.57 & 95.40 & 92.57 & 86.79 & 80.41 & 95.23 & 91.96 & 96.11 & \underline{97.09} & \textbf{97.19}  \\
		& $\rho_{1.05}\uparrow$ & 87.64 & 97.82 & 97.01 & 98.77 &  94.22 & 98.76 & 98.20 & 99.10 & \underline{99.70} & \textbf{99.72}  \\
		& $\rho_{1.10}\uparrow$ & 97.01 & 98.66 & 98.29 & 99.57 & 96.82 &  99.53 & 99.31 & 99.63 & \textbf{99.94} & \textbf{99.94}  \\ \midrule
		\multirow{5}{*}{\rotatebox[origin=l]{90}{\textbf{RUD-ToF}}}
		& MAE$\downarrow$ & 42.38 & 37.05 & 25.13 & 27.60 & 24.63 &  21.22 & 20.41 & 18.94 & \underline{17.29} & \textbf{16.73}  \\
		& RMSE$\downarrow$ & 121.61 & 71.36 & 61.50 & 49.94 & 43.68 & 48.76 & 33.83 & 31.78 & \underline{31.11} & \textbf{30.94}  \\
		& $\rho_{1.02}\uparrow$ & 63.99 & 48.04 & 66.23 & 61.65 & 56.04 & 62.03 & 67.30 & 68.41 & \textbf{70.13} & \underline{69.97}  \\
		& $\rho_{1.05}\uparrow$ & 84.71 & 80.40 & 87.33 & 81.57 & 79.25 & 87.04 & \underline{90.08} & 84.51 & 90.01  & \textbf{90.66} \\
		& $\rho_{1.10}\uparrow$ & 92.09 & 91.79 & 95.17 & 89.90 & 90.01 & 95.39 & \underline{96.91} & 94.92 & 96.74  & \textbf{96.97} \\ \bottomrule
	\end{tabular}\vspace{-0.3cm}
        \caption{\textbf{Comparison with the state-of-the-art on the SUD-ToF and RUD-ToF datasets.} The best and second-best results are marked in \textbf{bold} and \underline{underline}, respectively. The direction of arrows in metrics represents their trends (the lower/higher, the better).}\vspace{-0.2cm} 
  \label{sota_comparison}
\end{table*}

In NFC~\citet{nsampi2023neural}, the authors introduce a predefined Gaussian kernel with a continuous second derivative and set the control points to approximate the kernel using piecewise linear functions. 
When $n$ is set to $2$, the estimated kernel $\mathcal{\hat{K}}^{(2)}$ reduces to a sparse set of Dirac deltas $\delta$, which facilitates convolution calculations.
Different from it, which predefines the Gaussian kernel and control points, and then estimates Dirac deltas for convolution, our method directly generates estimated Dirac deltas $\mathcal{\hat{K}}^{(2)}$ through DISB. This design enhances the flexibility of kernel selection and reduces the complexity of the estimated procedure.
For the antiderivative computation $u^{(-2)}$ of the signal, a common approach is to use a trained coordinate-based neural network, typically structured as MLPs. However, the MLPs exhibit significant computational costs and are constrained to limited scenarios for training, inadequately addressing the demands of UD-ToF imaging. In contrast, we propose a simple yet efficient formulation of repeated antiderivatives in the continuous convolution based on Eq.(\ref{eq_convolution}). Our repeated antiderivatives are estimated as:
\begin{equation}
    u^{(-2)} \approx Au(x_0,y_0)+B
\end{equation}
where $A$ and $B$ are coefficients, and $(x_0,y_0)\in \mathbf{z}$ is the pixel where the continuous convolution is to be performed. The detailed proof of this process is given as:
\begin{equation}		
		\resizebox{0.9\hsize}{!}{$\begin{aligned}
            &\int\int u\text{d}\mathbf{z} 
            \approx \sum_{i=0}^{m} \sum_{j=0}^{n} u(x_i,y_j)\cdot \Delta \mathbf{z} \\
            &=u(x_0,y_0)\cdot \Delta \mathbf{z}+u(x_1,y_0)\cdot \Delta \mathbf{z} +\cdots + u(x_m,y_n)\cdot \Delta \mathbf{z}\\
            &=[u(x_0,y_0)+C_{0,0}]\cdot \Delta \mathbf{z}+[u(x_0,y_0)+C_{1,0}]\cdot \Delta \mathbf{z}+ \cdots \\
            &\quad  +[u(x_0,y_0)+C_{m,n}]\cdot \Delta \mathbf{z} \\
            &=\underbrace{(m+1)(n+1) \Delta \mathbf{z} }_{A}\cdot u(x_0,y_0)+\underbrace{\sum_{i=0}^{m} \sum_{j=0}^{n} C_{i,j} \cdot \Delta \mathbf{z}}_{B} 
		\end{aligned}$}
\end{equation}
where $C_{0,0}=0$. As shown in Fig. \ref{fig:CC2d}, the deep initial state builder additionally outputs amplitudes $I_A$, features $F$, and offsets $O_f$ before iteration. During iteration, these are concatenated and processed through three convolution layers, with the middle layer having only $32$ channels, to yield the coefficients $A$ and $B$.  

Although both our proposed continuous convolution and NFC~\cite{nsampi2023neural} are inspired by repeated differentiation, there are fundamental differences in the process, from the generation of Dirac deltas to the computation of signal antiderivatives.

\begin{figure*}[t] 
	\centering
	\renewcommand\tabcolsep{1.5pt} 
	\begin{tabular}{ccccccccccc}
	\vspace{-0.16cm}
        \includegraphics[width=0.75in]{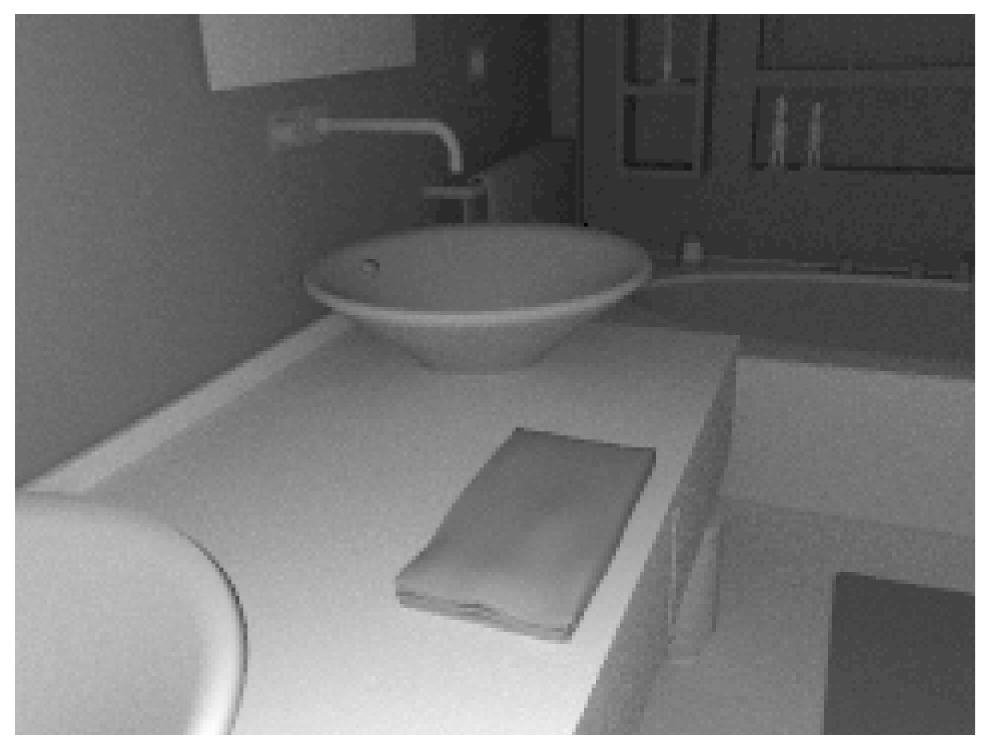}
	\hspace{-2.3mm} & \includegraphics[width=0.75in]{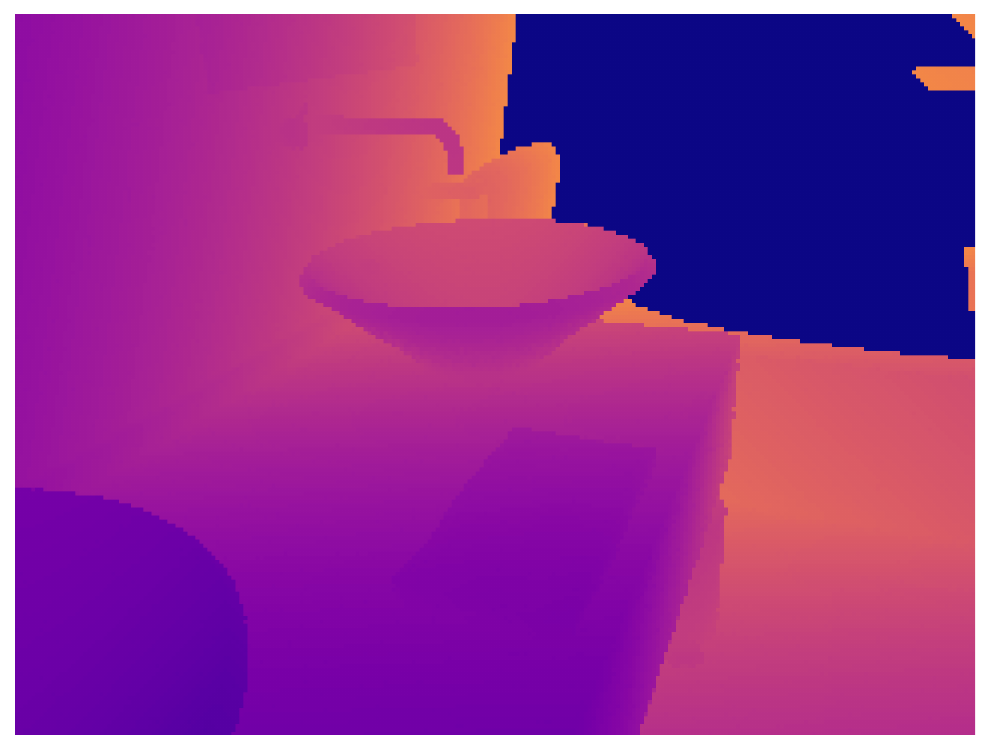}
	\hspace{-2.3mm} & \includegraphics[width=0.75in]{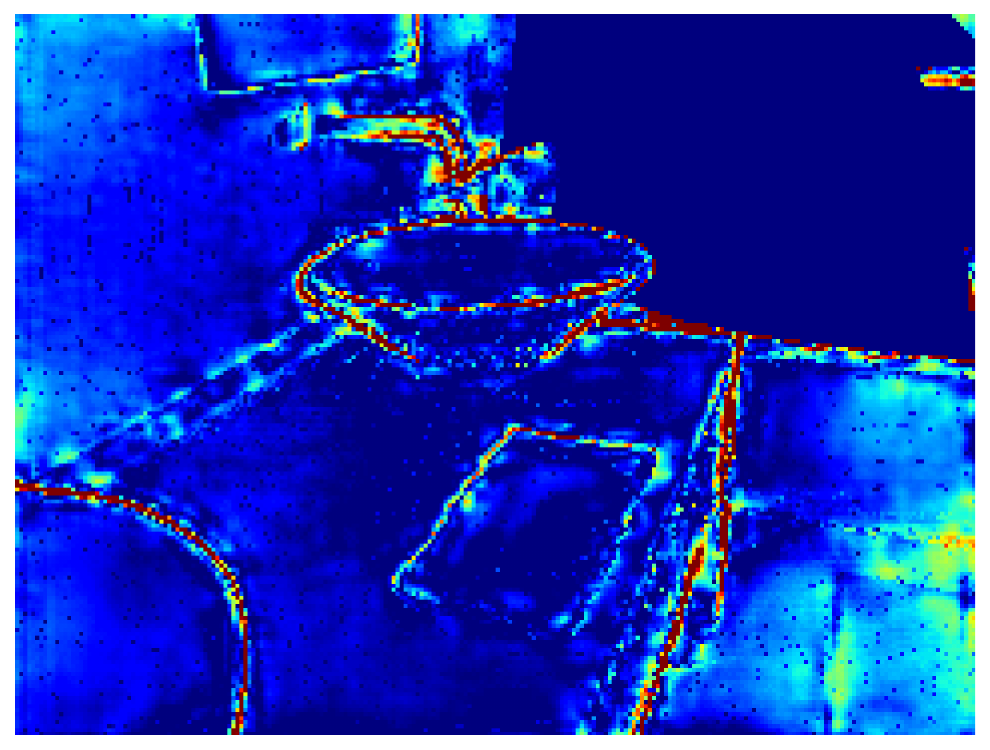}
	\hspace{-2.3mm} & \includegraphics[width=0.75in]{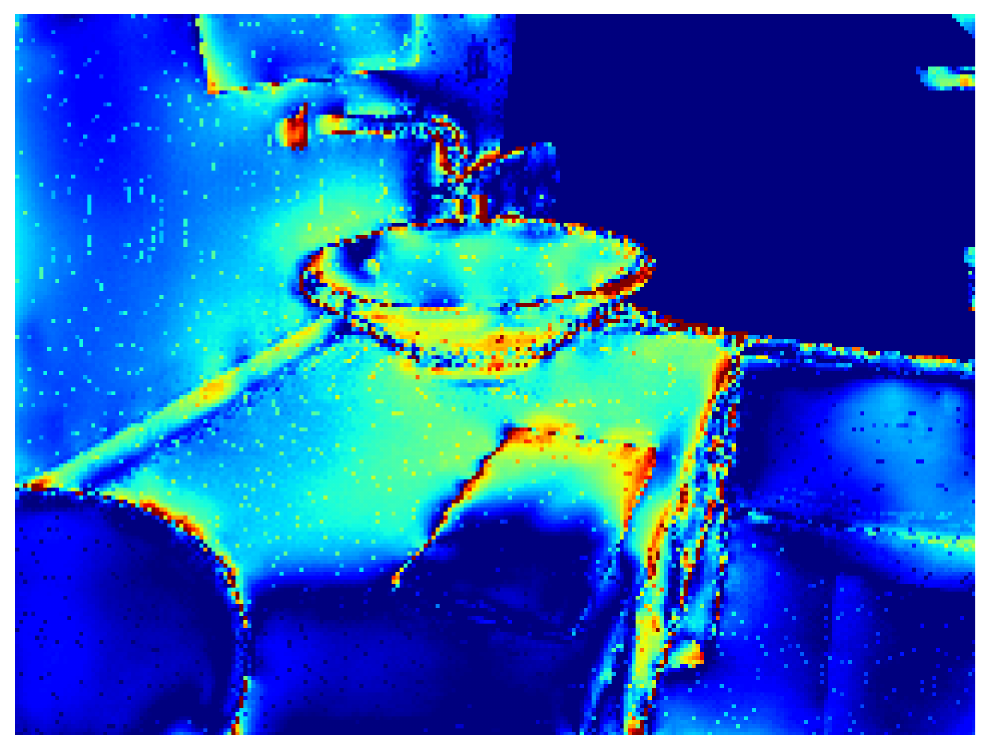}
	\hspace{-2.3mm} & \includegraphics[width=0.75in]{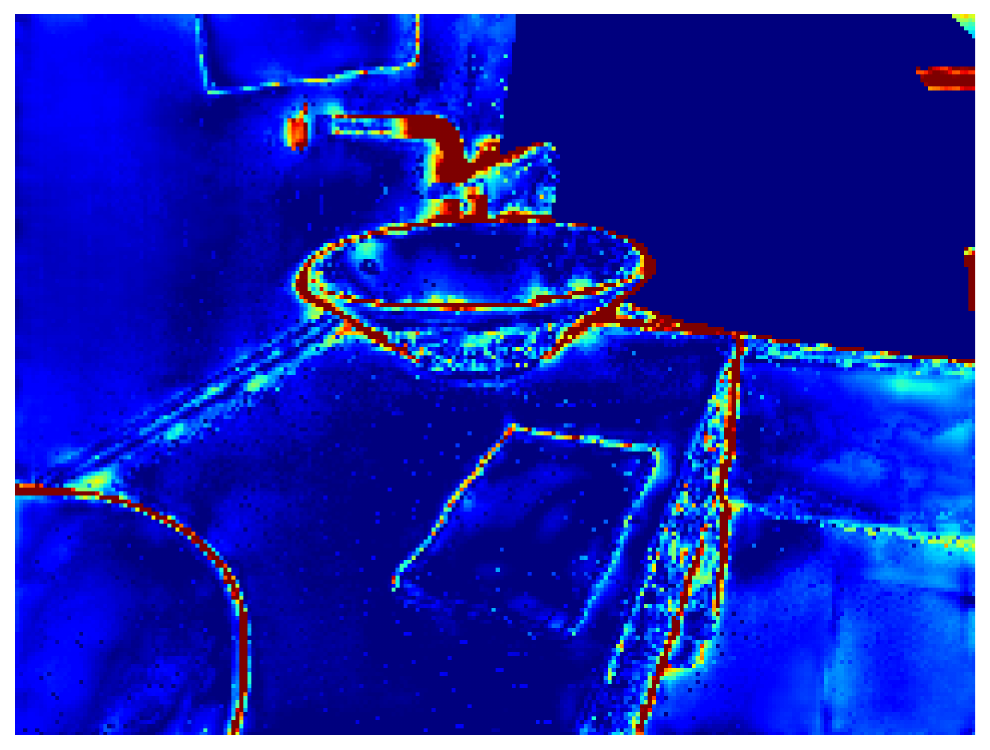}
	\hspace{-2.3mm} & \includegraphics[width=0.75in]{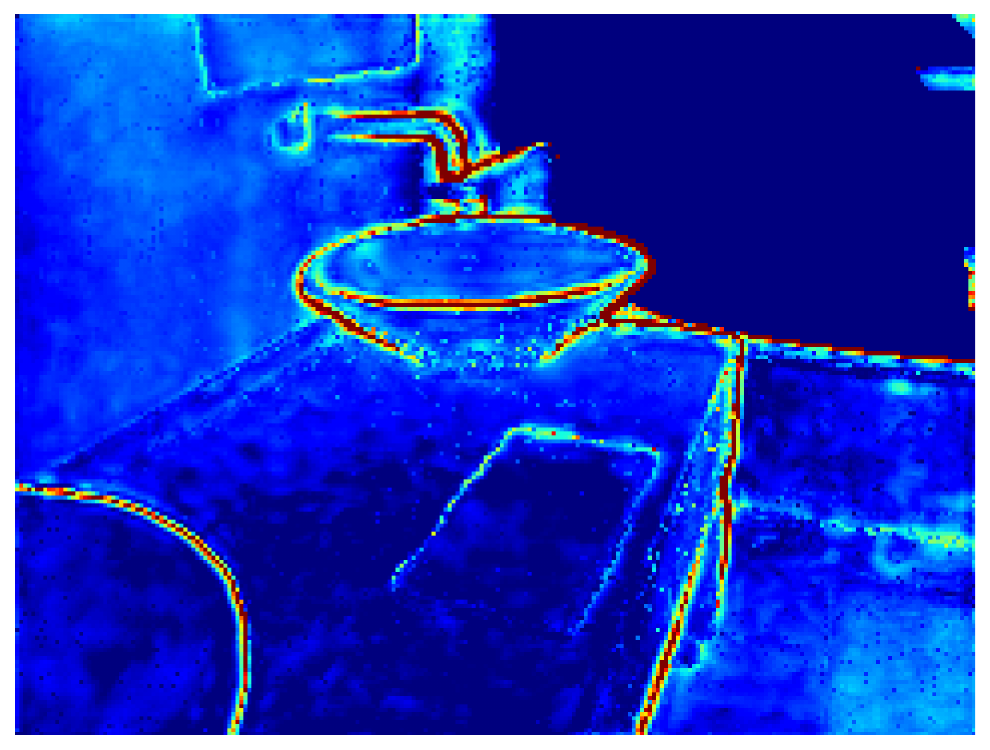}
	\hspace{-2.3mm} & \includegraphics[width=0.75in]{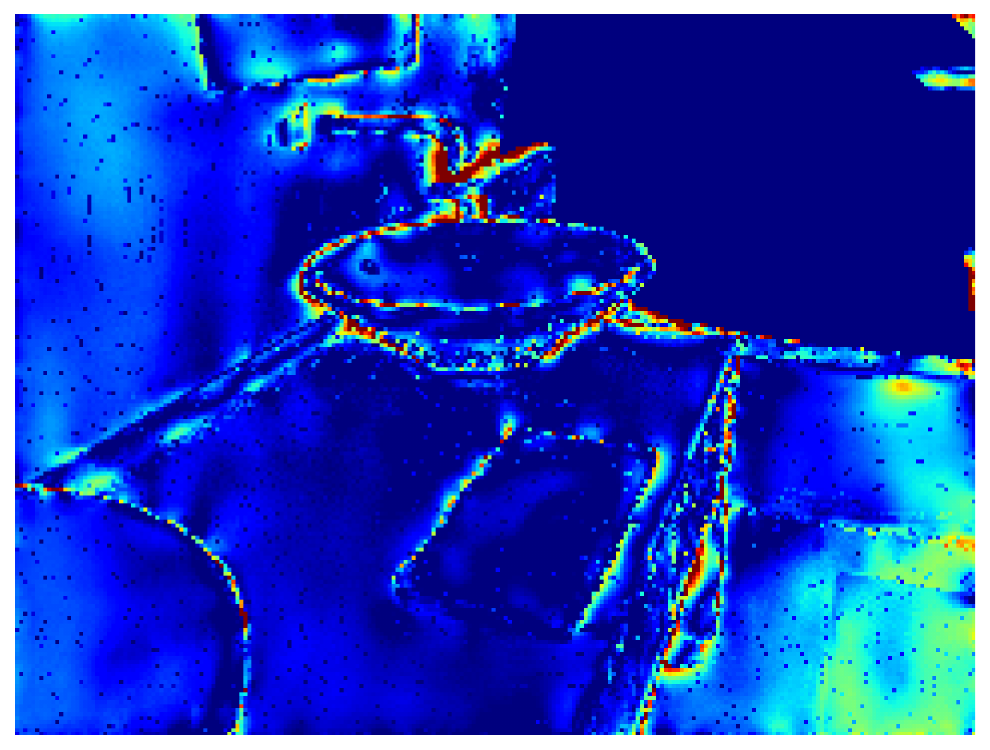}
	\hspace{-2.3mm} & \includegraphics[width=0.75in]{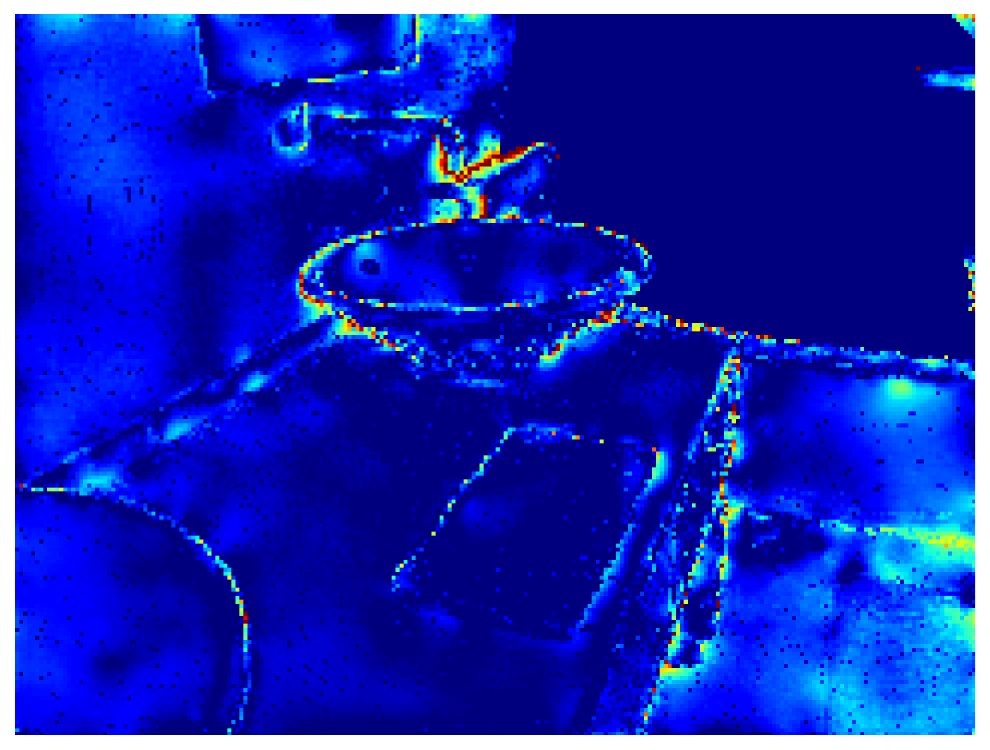}
 
	\hspace{-2.3mm} & \includegraphics[width=0.75in]{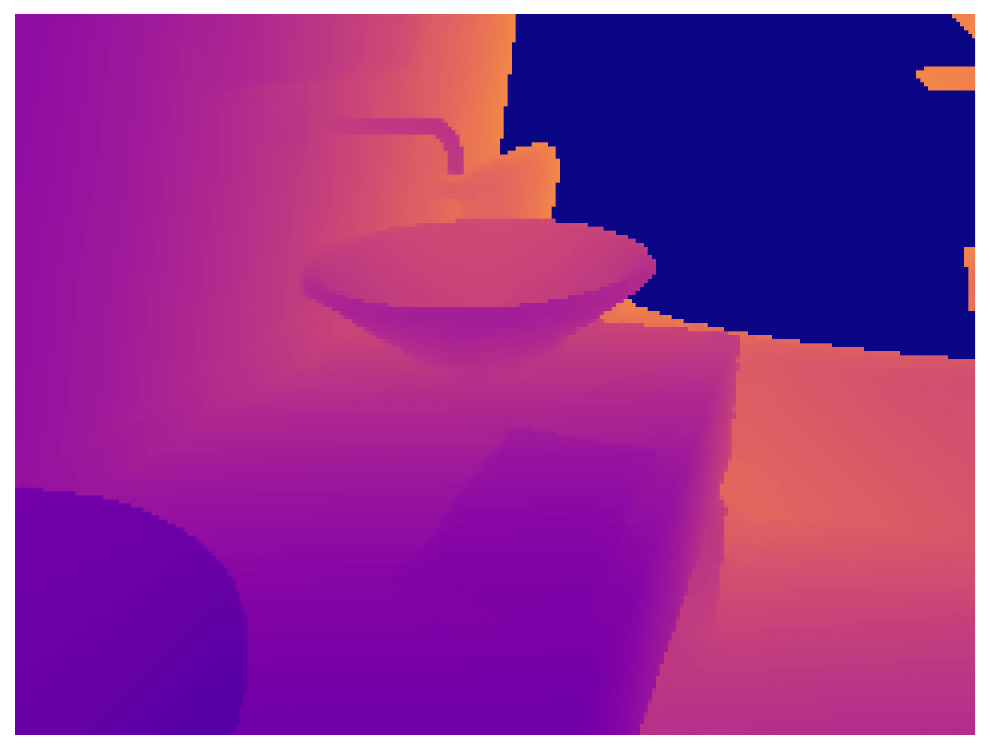}
        \\ \vspace{-0.6mm}
        
        \includegraphics[width=0.75in]{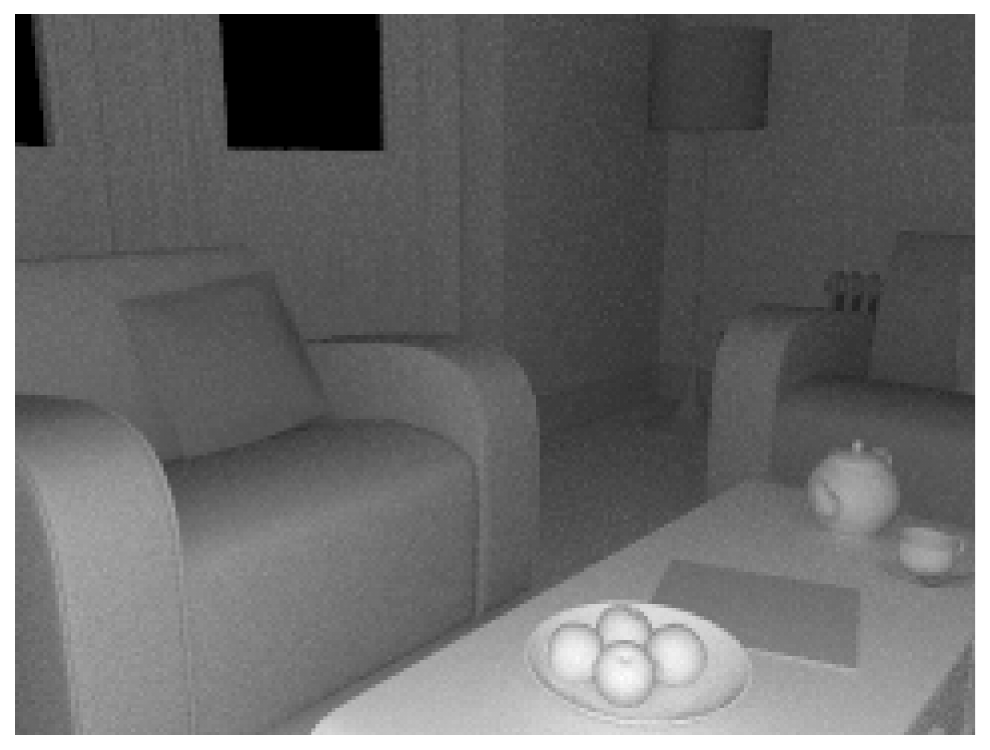}
	\hspace{-2.3mm} & \includegraphics[width=0.75in]{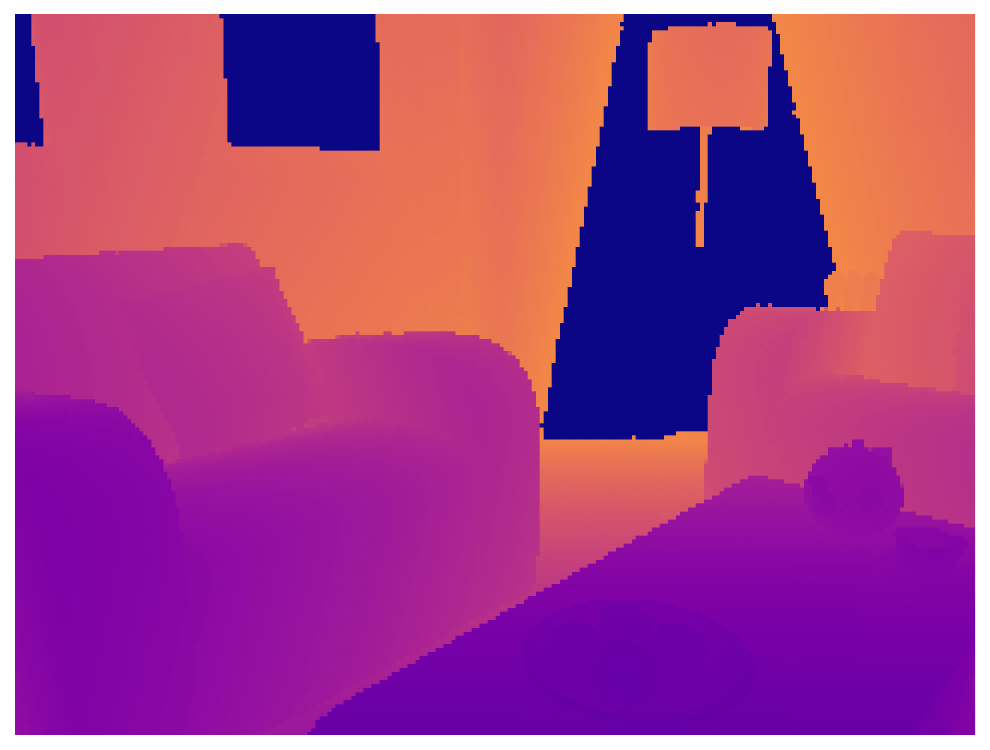}
	\hspace{-2.3mm} & \includegraphics[width=0.75in]{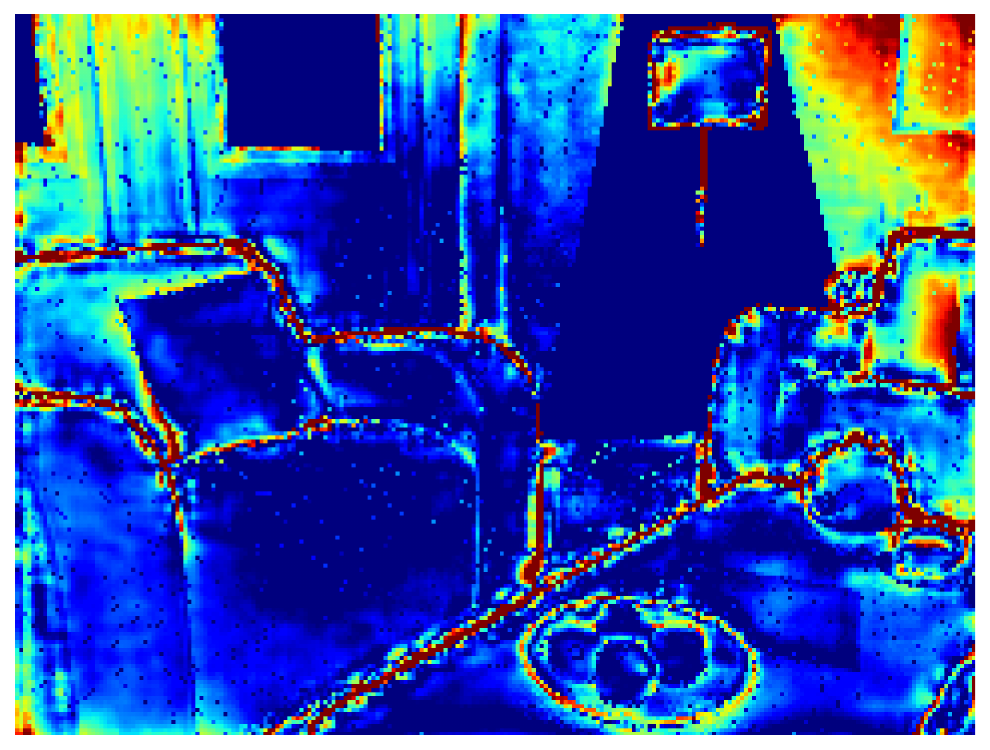}
	\hspace{-2.3mm} & \includegraphics[width=0.75in]{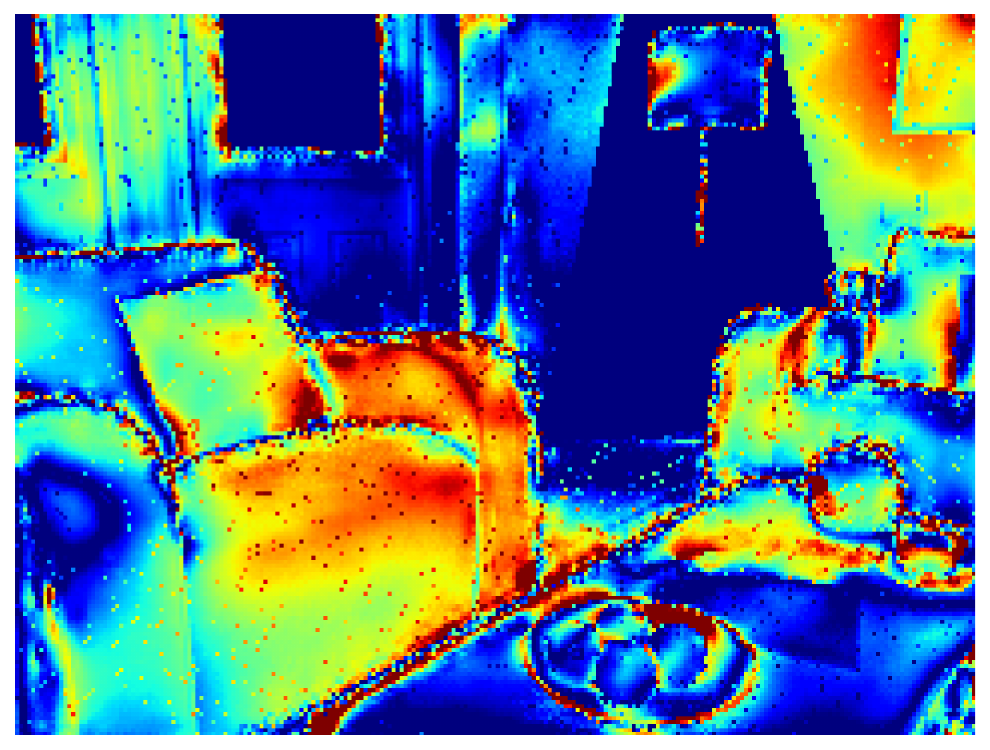}
	\hspace{-2.3mm} & \includegraphics[width=0.75in]{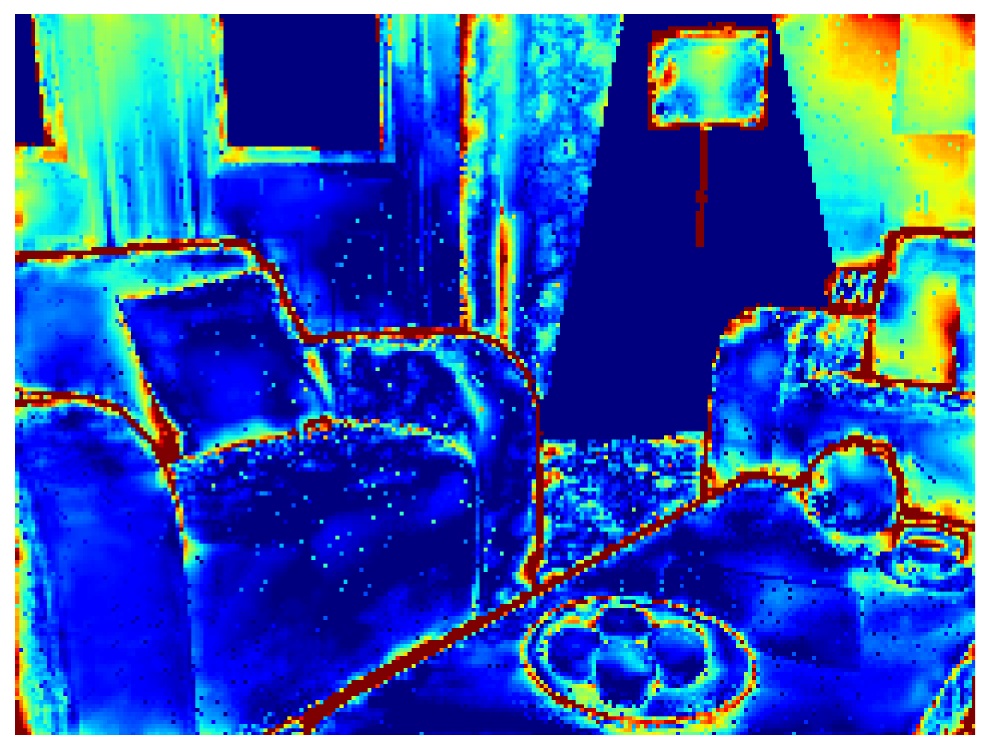}
	\hspace{-2.3mm} & \includegraphics[width=0.75in]{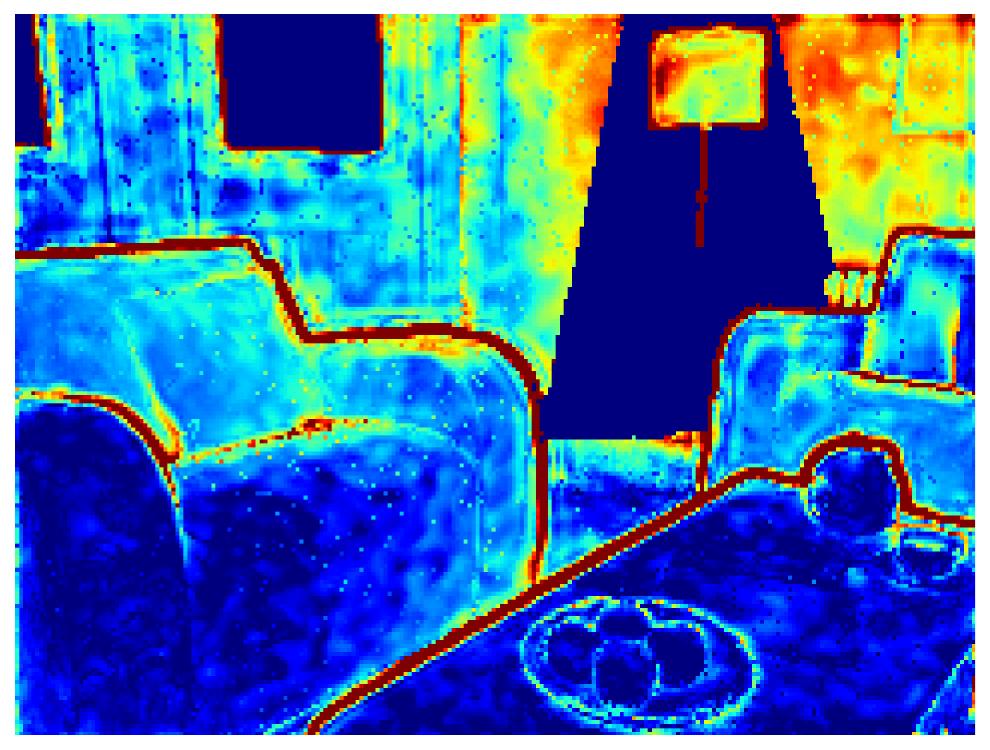}
	\hspace{-2.3mm} & \includegraphics[width=0.75in]{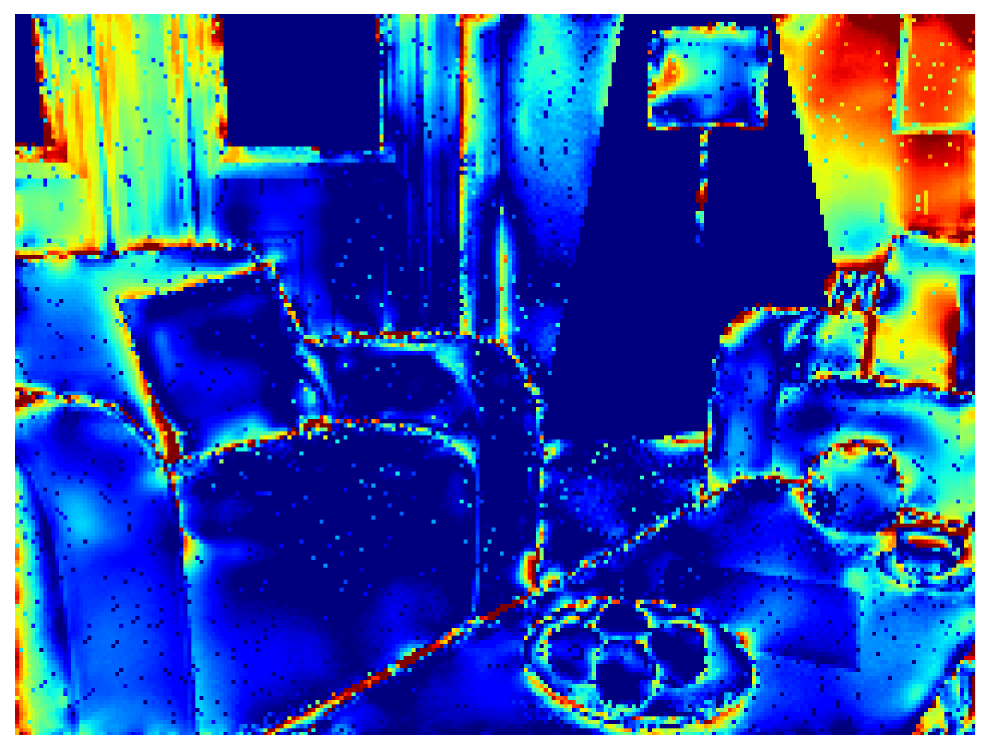}
	\hspace{-2.3mm} & \includegraphics[width=0.75in]{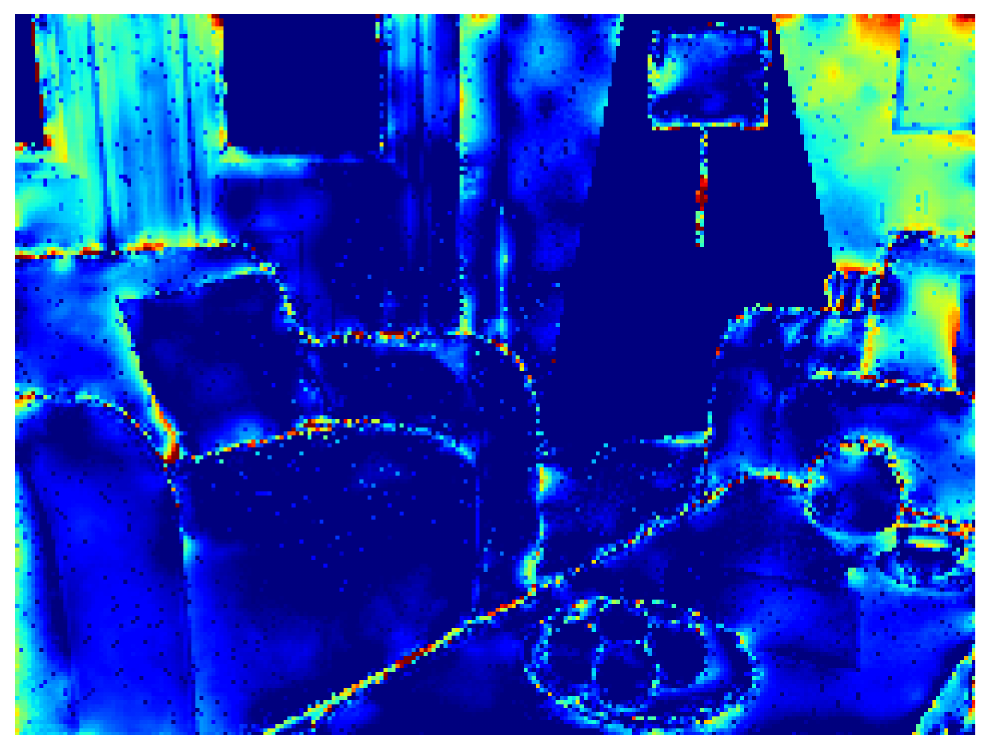}
 
	\hspace{-2.3mm} & \includegraphics[width=0.75in]{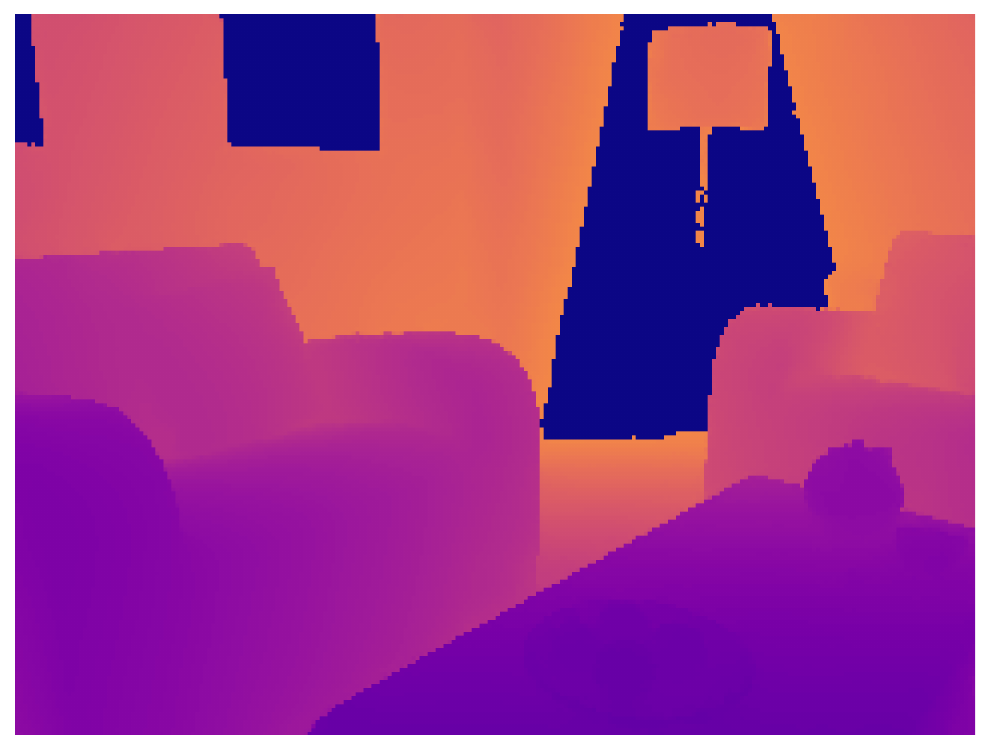}
         \\ \vspace{-0.16cm}

        \includegraphics[width=0.75in]{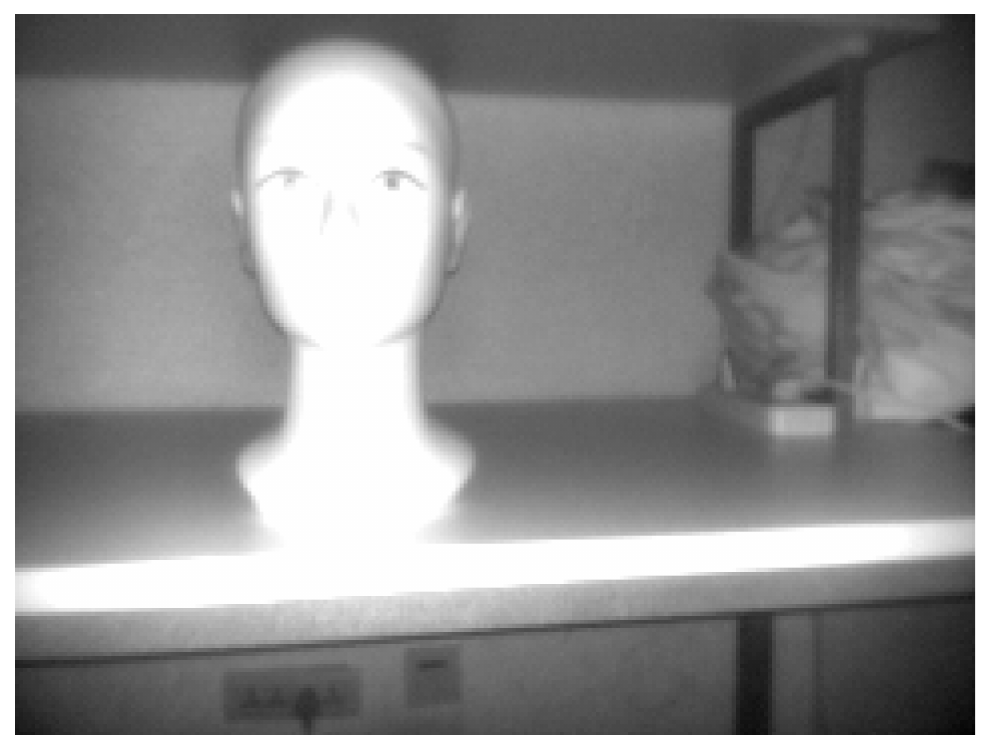}
        \hspace{-2.3mm} & \includegraphics[width=0.75in]{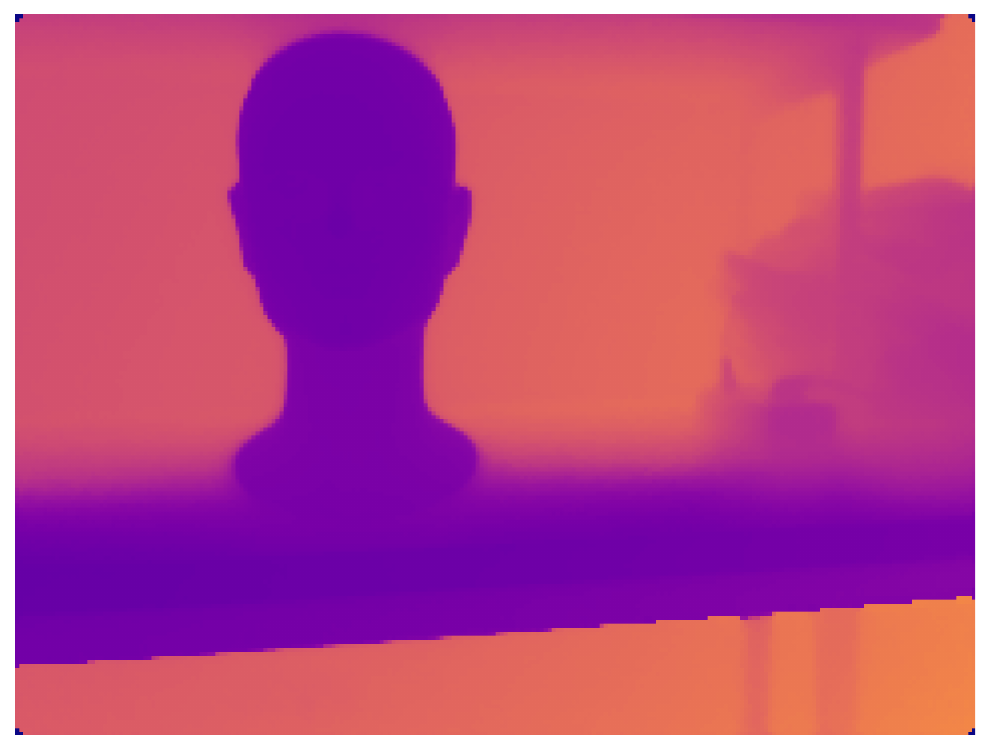}
	\hspace{-2.3mm} &  \includegraphics[width=0.75in]{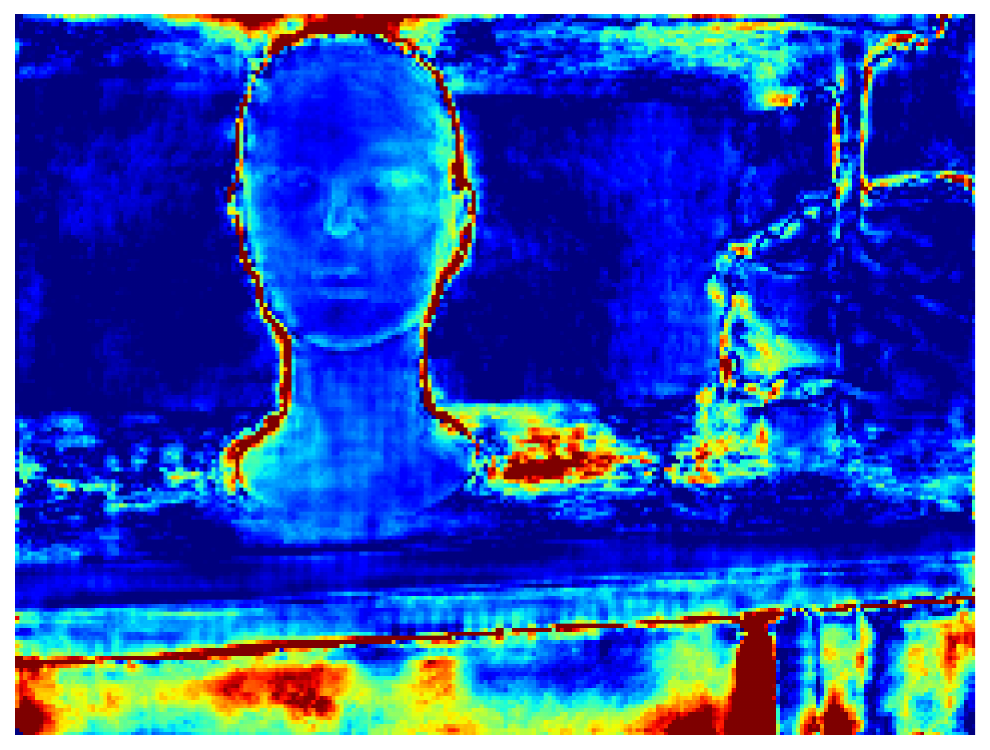}
	\hspace{-2.3mm} & \includegraphics[width=0.75in]{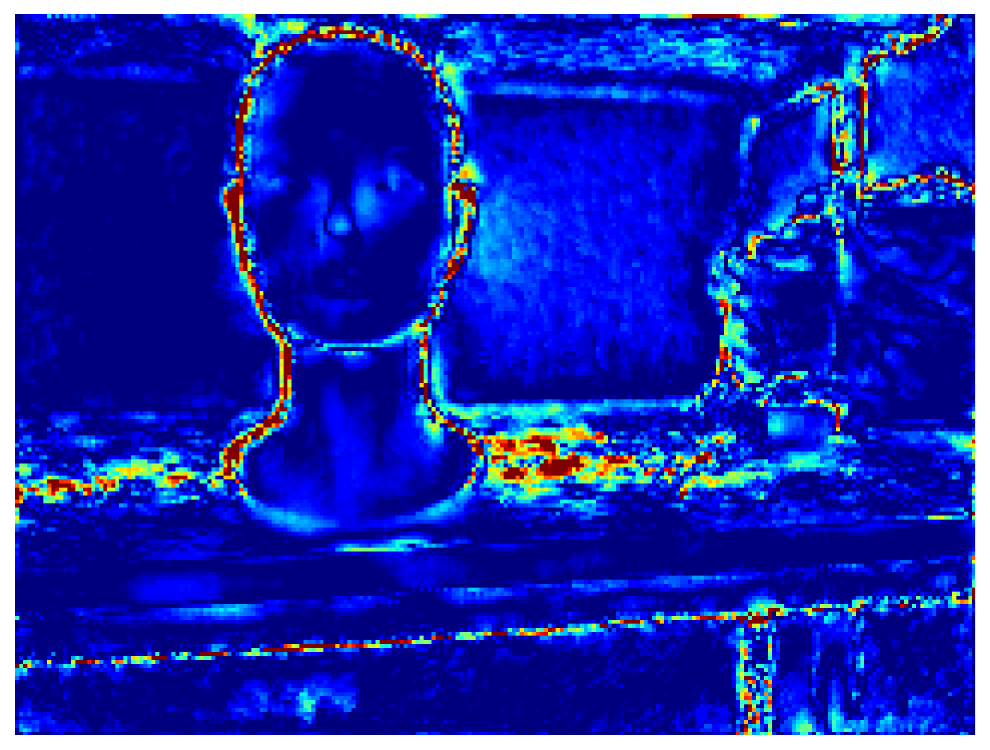}
	\hspace{-2.3mm} & \includegraphics[width=0.75in]{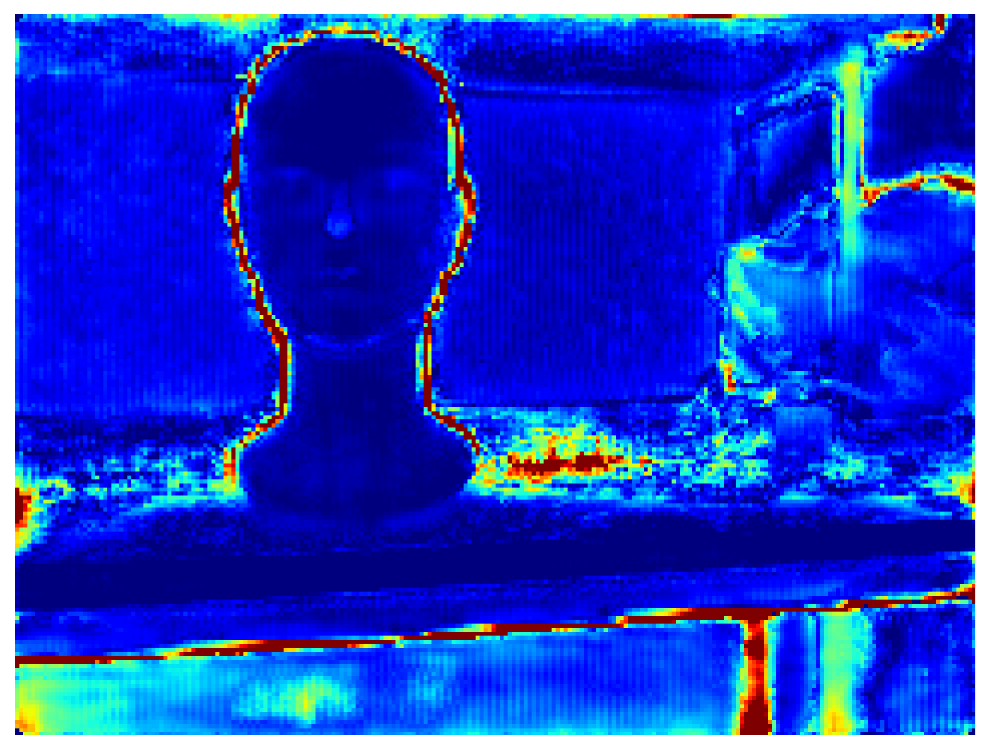}
	\hspace{-2.3mm} & \includegraphics[width=0.75in]{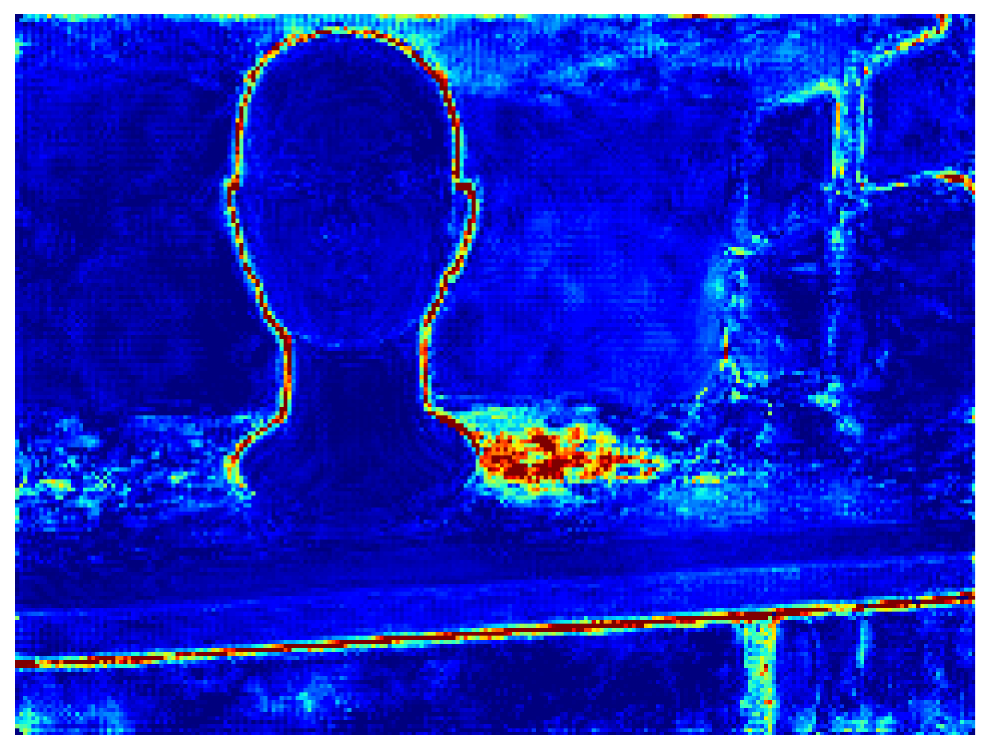}
	\hspace{-2.3mm} & \includegraphics[width=0.75in]{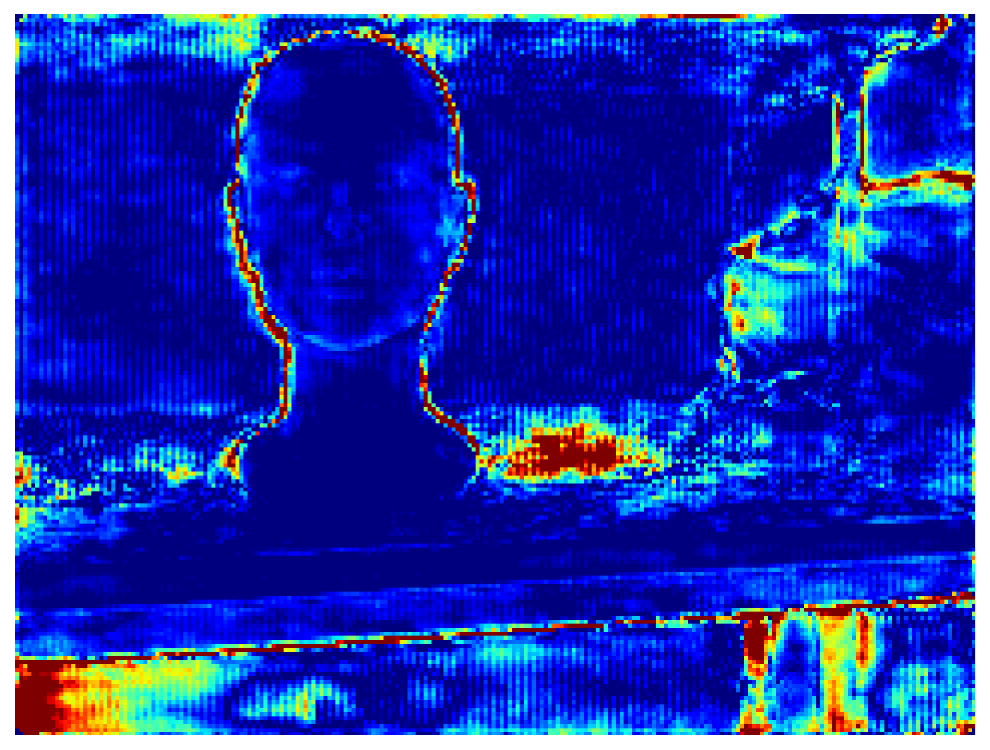}
	\hspace{-2.3mm} & \includegraphics[width=0.75in]{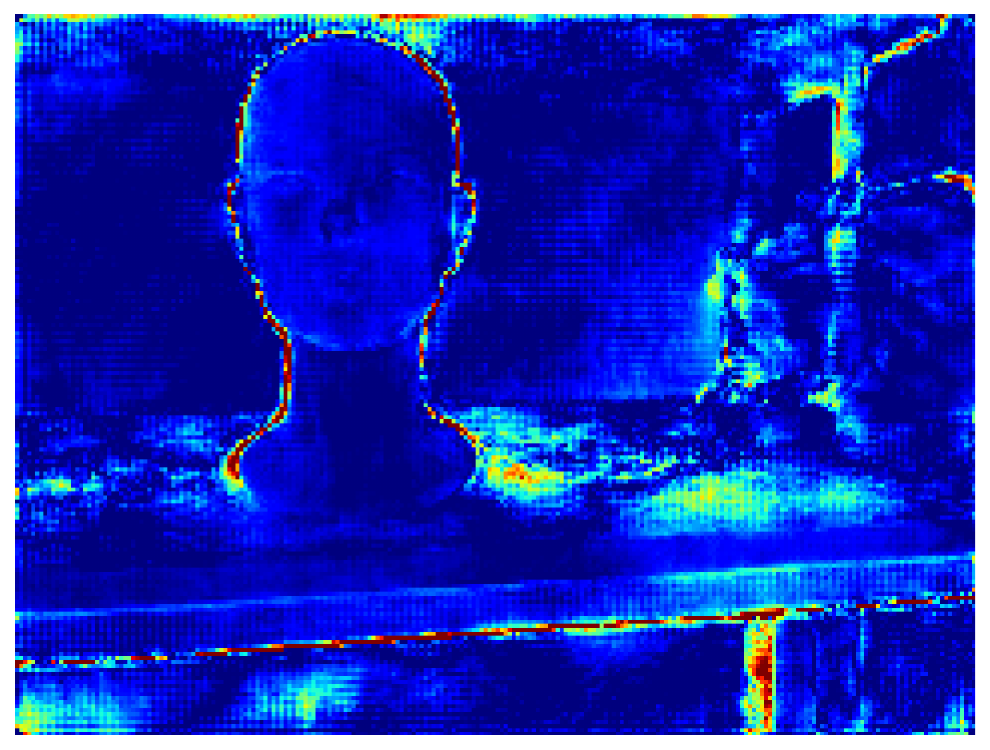}
        
        \hspace{-2.3mm} & \includegraphics[width=0.75in]{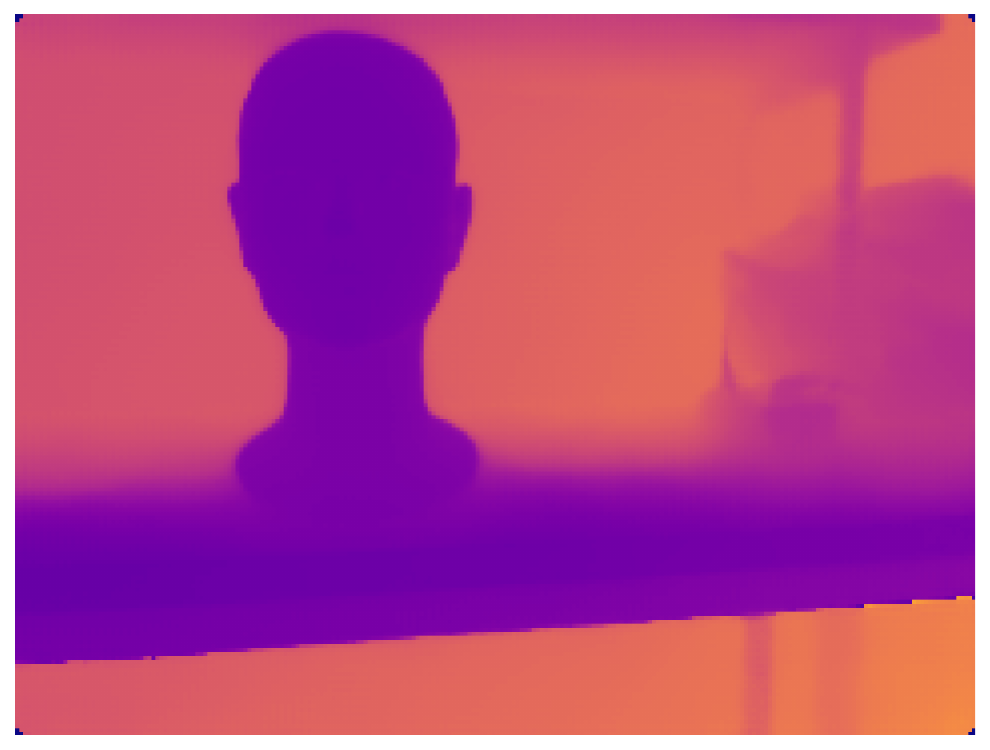}
    \\ \vspace{-0.1cm}
    
        \includegraphics[width=0.75in]{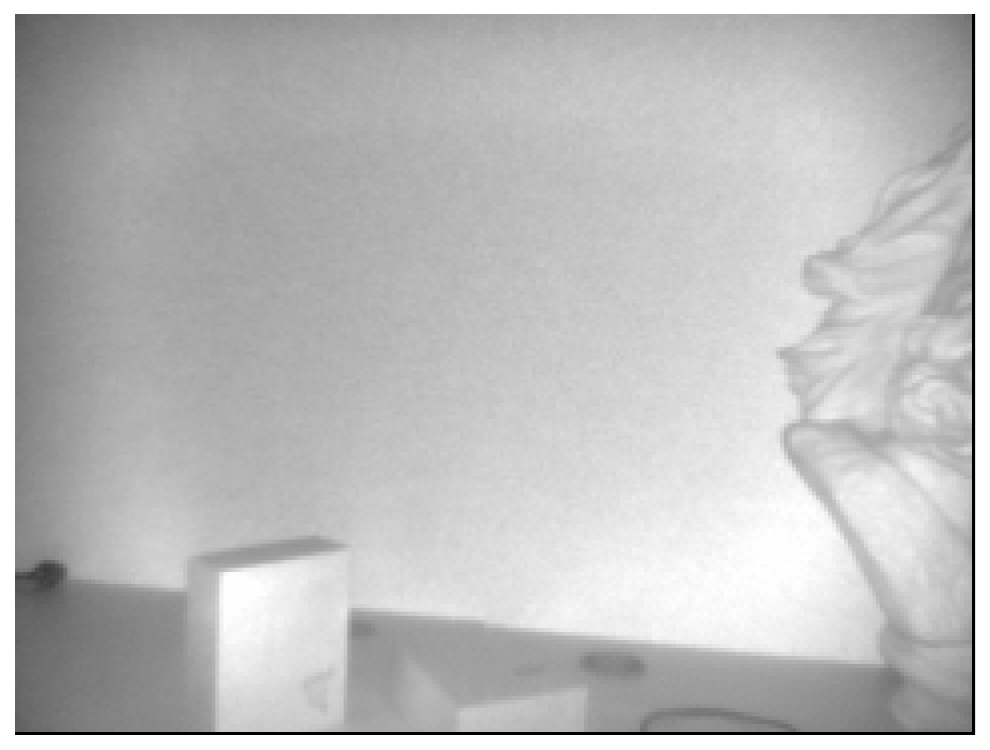}
        \hspace{-2.3mm} & \includegraphics[width=0.75in]{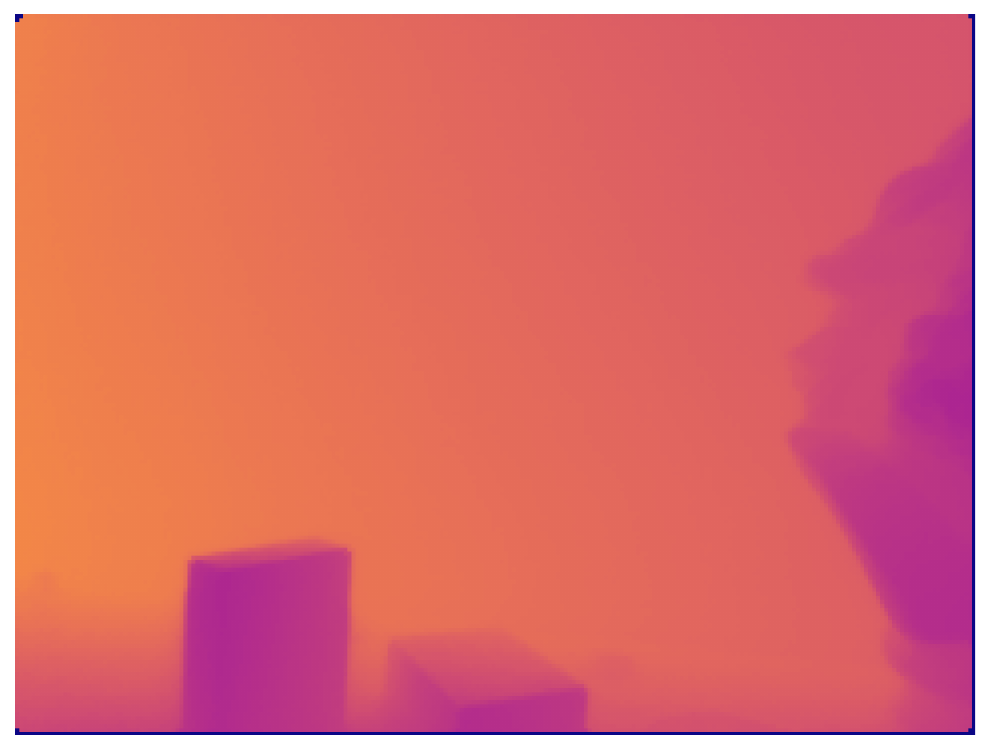}
	\hspace{-2.3mm} &  \includegraphics[width=0.75in]{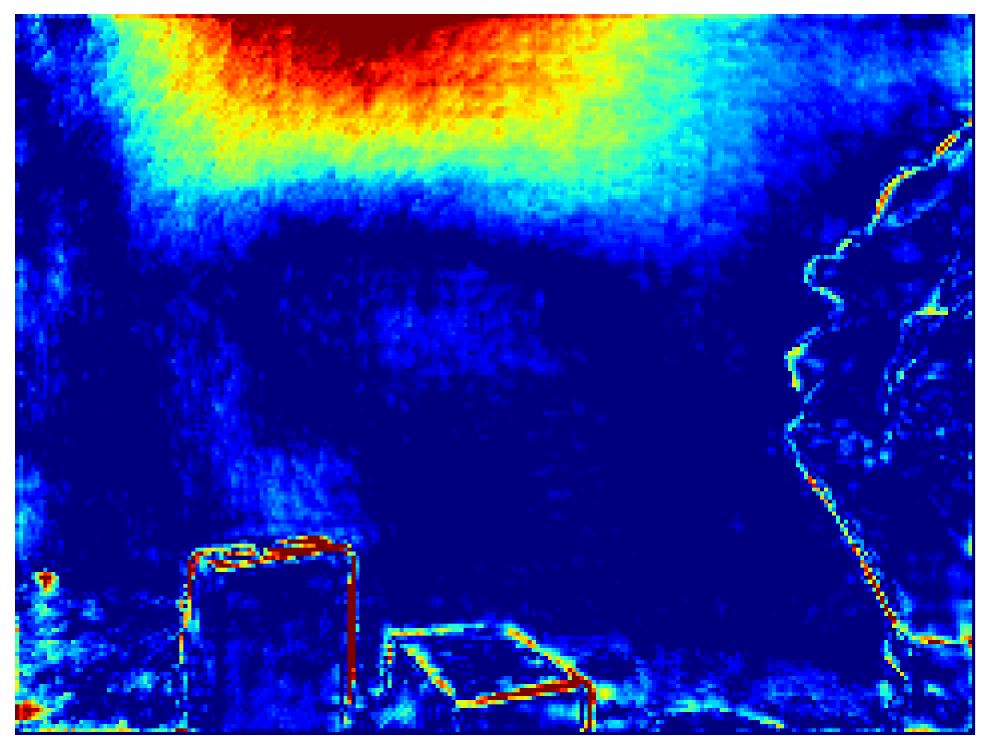}
	\hspace{-2.3mm} & \includegraphics[width=0.75in]{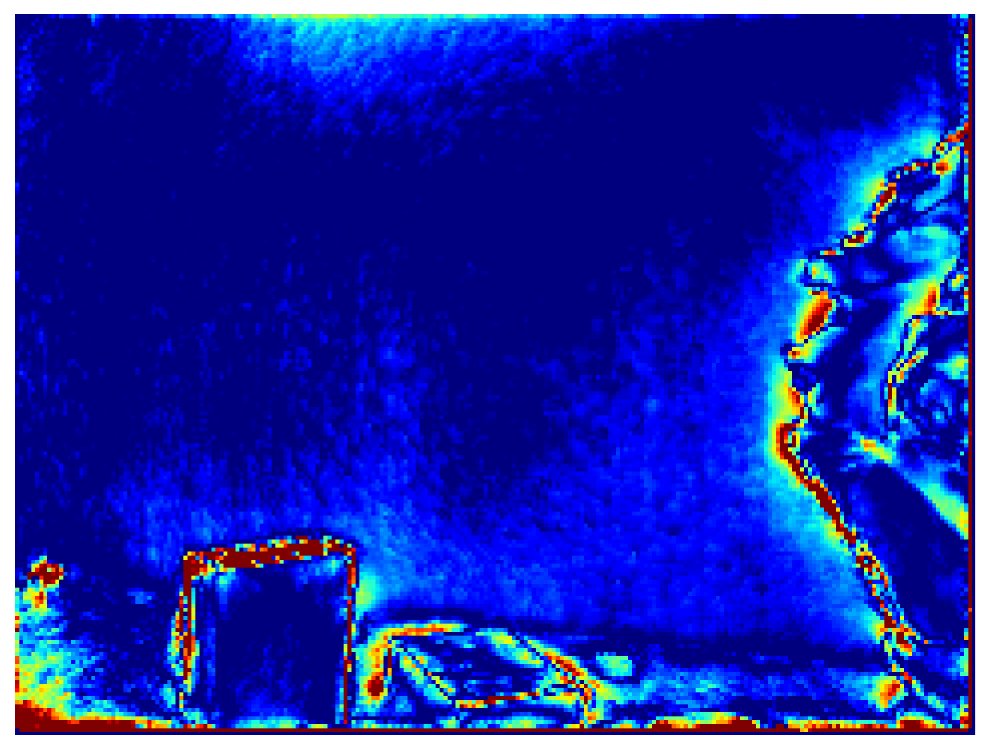}
	\hspace{-2.3mm} & \includegraphics[width=0.75in]{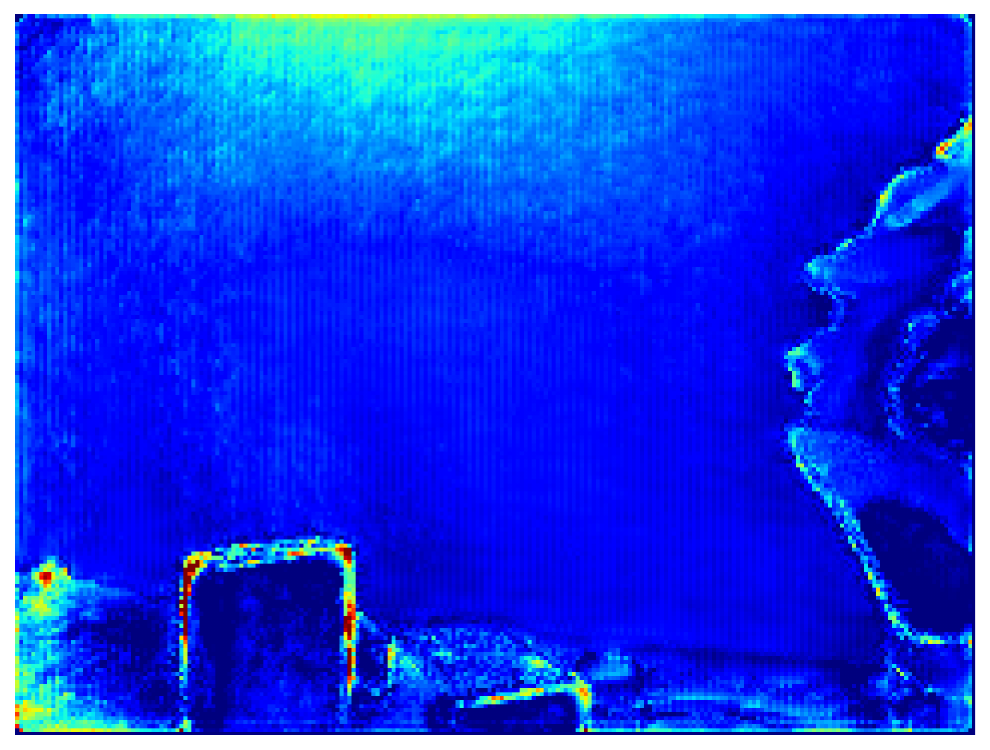}
	\hspace{-2.3mm} & \includegraphics[width=0.75in]{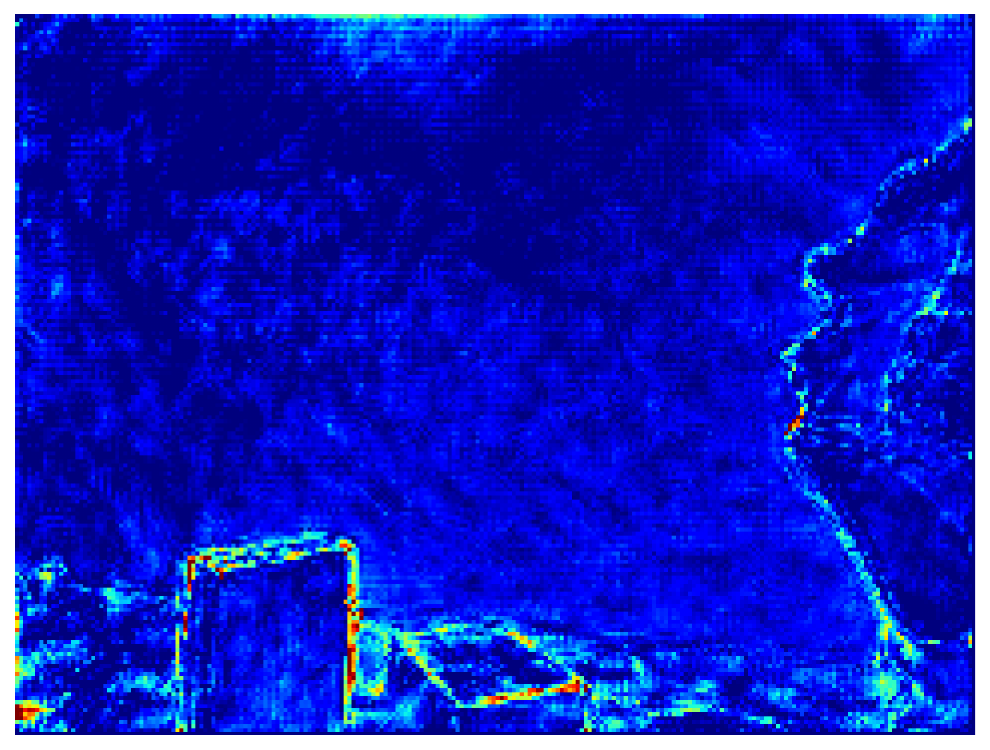}
	\hspace{-2.3mm} & \includegraphics[width=0.75in]{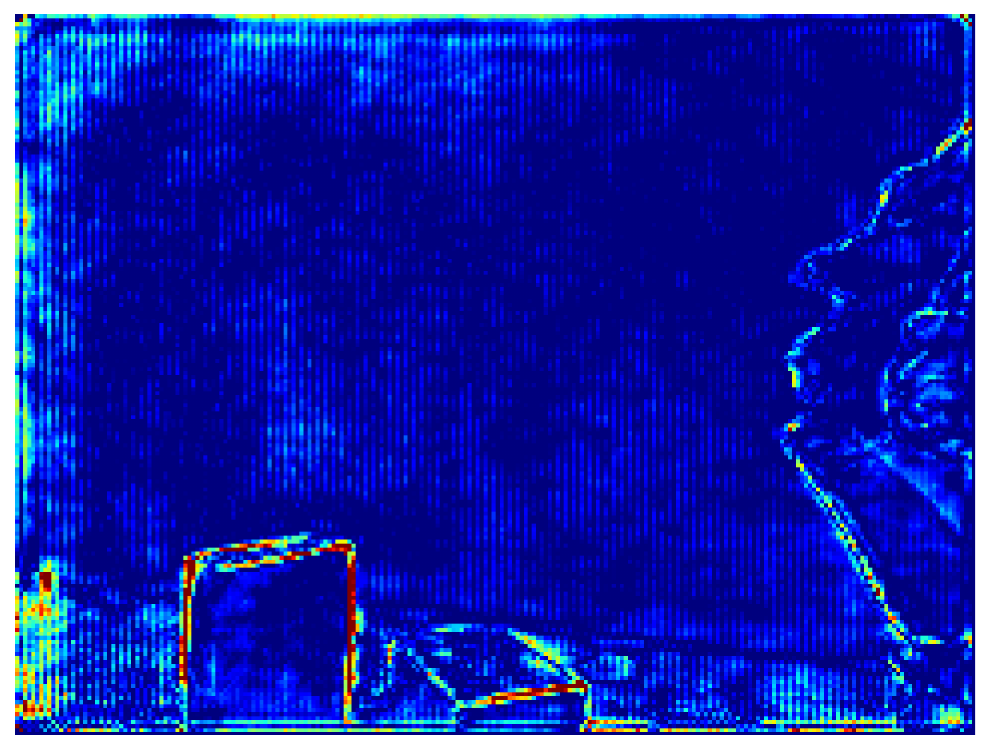}
	\hspace{-2.3mm} & \includegraphics[width=0.75in]{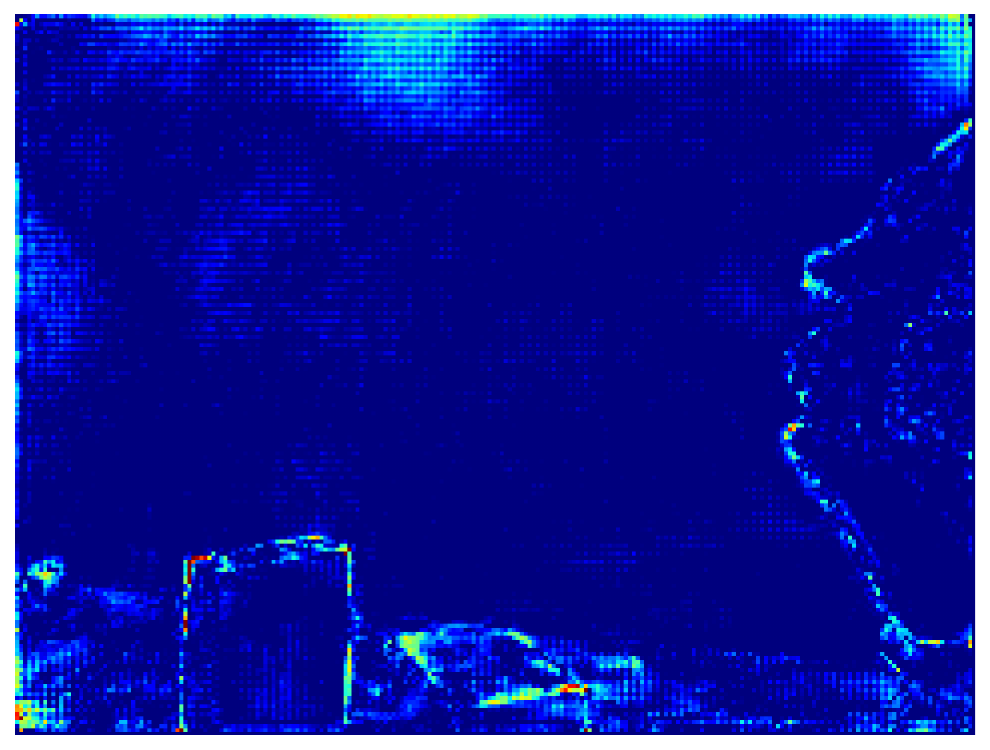}
        
        \hspace{-2.3mm} & \includegraphics[width=0.75in]{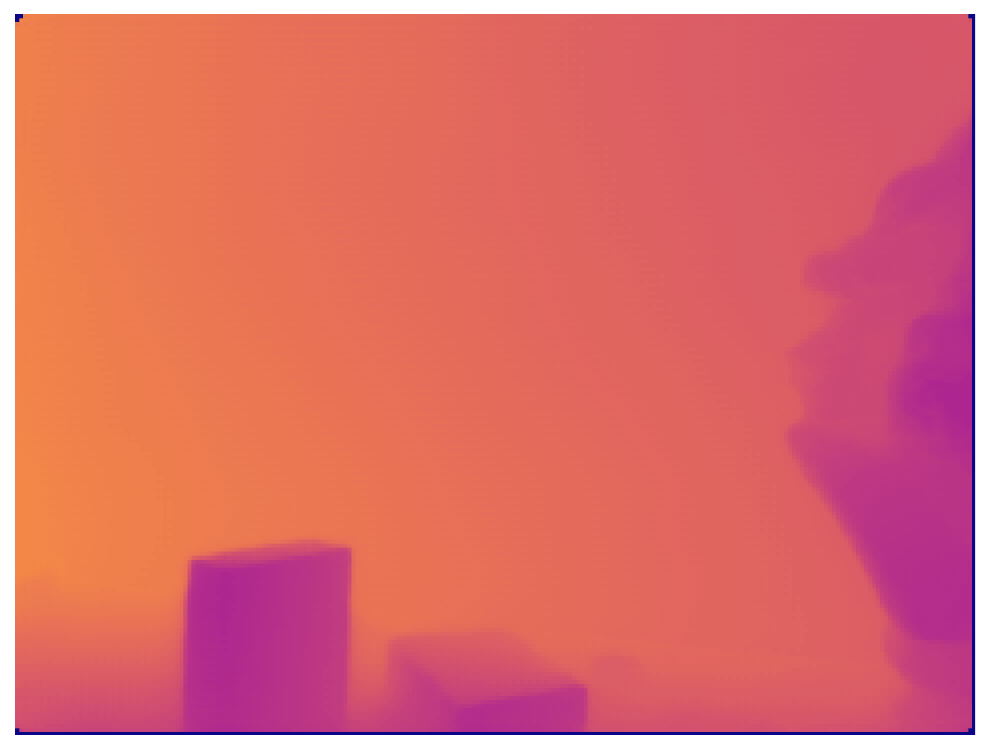}
 \\
    {\fontsize{8pt}{8pt}\selectfont \textbf{(a)} IR} & {\fontsize{8pt}{8pt}\selectfont \textbf{(b)} GT} & {\fontsize{8pt}{8pt}\selectfont \textbf{(c)} PE-ToF} & {\fontsize{8pt}{8pt}\selectfont \textbf{(d)} SHARPnet} & {\fontsize{8pt}{8pt}\selectfont \textbf{(e)} Restormer} & {\fontsize{8pt}{8pt}\selectfont \textbf{(f)} NAFNet} & {\fontsize{8pt}{8pt}\selectfont \textbf{(g)} UD-ToFnet} & {\fontsize{8pt}{8pt}\selectfont \textbf{(h)} Ours} & {\fontsize{8pt}{8pt}\selectfont \textbf{(i)} Our Pred}
	\end{tabular}
    \vspace{-0.3cm}
	\caption{\textbf{Qualitative results on the SUD-ToF (top two rows) and RUD-ToF (bottom two rows) dataset.} From left to right: (a) IR image and (b) ground-truth depth, followed by (c-g) error maps achieved by SoTA solutions and (h) LFRD$^2$, (i) depth maps by LFRD$^2$. }\vspace{-0.2cm} 
	\label{fig:sota_comp1}
\end{figure*}

\begin{table*}[htbp] \small   	
	\renewcommand\tabcolsep{5.5pt} 
	\centering
	\begin{tabular}{@{}ccccccccccc@{}}
		\toprule
		\multicolumn{1}{c}{\textbf{Dataset}} & \multicolumn{1}{c}{\textbf{Metrics}} & CDNLM & JGDR & ToFnet & ToF-KPN & SHARPnet & Cardioid & PE-ToF & UD-ToFnet & LFRD$^2$ \\ \midrule
		\multirow{2}{*}{FLAT}
		& MAE$\downarrow$ & 13.86 & 8.86 & 54.33 & 4.65 & 4.62 & 6.74 & 7.93 & \underline{4.41}  & \textbf{4.13} \\
		& RMSE$\downarrow$ & 21.00 & 45.78 & 74.99 & 12.83 & 10.26 & 19.94 & 32.60 & \underline{8.23}  & \textbf{7.35} \\
        \bottomrule
	\end{tabular}\vspace{-0.3cm}
        \caption{\textbf{Comparison with the state-of-the-art on the FLAT dataset.} The best and second best results are marked in \textbf{bold} and \underline{underline}, respectively. The direction of arrows in metrics represents their trends (the lower/higher, the better).}\vspace{-0.2cm} 
        \label{sota_flat}
\end{table*}


\section{Experiments}

We now report the outcome of our evaluation. We first introduce the experimental settings, then we compare LFRD$^2$ with state-of-the-art solutions in UD-ToF imaging, conduct ablation studies, and conclude by discussing limitations.

\subsection{Experimental Settings}
We use two public UD-ToF datasets to evaluate the effectiveness of our proposal: SUD-ToF and RUD-ToF \cite{qiao2022depth}. 
The former is a synthetic dataset crafted via transient rendering, encompassing 100K images, of which $10\%$ are randomly allocated for testing. The latter is a real UD-ToF dataset featuring diverse indoor scenes, divided into 1171 scenes for training and 105 for evaluation.
MAE (Mean Absolute Error) and RMSE (Root Mean Squared Error), both measured in millimeters (mm), are used as evaluation metrics.
Additionally, we measure the proportion (denoted as $\rho_{th}$) of pixels that fall within a specified relative error range compared to the total number of pixels. Following \citet{qiao2022depth}, we set $th\in\{1.02,1.05,1.10\}$.
To facilitate the network processing, the original image of $180\times 240$ is cropped to a patch of $176\times240$.
We use the Adam optimizer and a batch size of 16. The initial learning rate is set to $1\times 10^{-4}$. The total number of epochs is 250 for SUD-ToF and 1000 for RUD-ToF. 
The proposed method is implemented using the Pytorch framework and experiments are conducted on a Nvidia RTX 3090 GPU. 
Moreover, the deep initial state builder is based on UD-ToFnet \cite{qiao2022depth}, keeping the original settings.

To prove the effectiveness of our framework beyond UD-ToF imaging, we conduct experiments on two additional datasets.
FLAT \cite{guo2018tackling} is a synthetic dataset of 2000 ToF measurements, capturing several nonidealities affecting real ToF sensors; we use it to perform ToF depth map denoising.
NYUv2 \cite{Silberman:ECCV12} comprises video sequences from various indoor scenes, collected by both the RGB and Depth cameras from the Microsoft Kinect for a total of 1449 RGBD frames; we deploy this dataset to perform depth super-resolution, following standard settings from the literature \cite{qiao2023depth}.

\begin{table}[t] \small
	\renewcommand\tabcolsep{10pt} 
	\centering
	\begin{tabular}{@{}lccc@{}}
		\toprule
            \textbf{Methods} & $4\times$ & $8\times$ & $16\times$    \\
            \midrule
		MSG         &  6.85 / 0.81 & 24.1 / 1.66 & 84.5 / 3.35   \\   
		FDKN   & 9.07 / 0.85 & 29.9 / 1.80 & 113 / 3.95   \\            
		PMBANet    & 10.8 / 0.93 & 17.2 / 1.38 & 84.9 / 3.26     \\         
		FDSR       & 10.1 / 0.94 & 19.5 / 1.38 & 86.4 / 3.35     \\         
		DCTNet  & 3.63 / 0.68 & 20.9 / 1.79 & 77.0 / 3.61    \\          
		LGR         & 6.45 / 0.73 & 19.6 / 1.42 & 67.5 / 2.90 \\           
		DADA    & 4.83 / 0.64 & 16.6 / 1.30 & 59.0 / 2.64    \\        
        SGNet	          &  3.22 / 0.54 & 14.9 / 1.26   & 58.8 / \underline{2.63}  \\   
        DSR-EI           & \underline{2.94} / \underline{0.49} & \underline{13.3} / \underline{1.19} &  \underline{57.0} / 2.70   \\   
        LFRD$^2$                          &  \textbf{2.85} / \textbf{0.47} & \textbf{12.8} / \textbf{1.16} &  \textbf{52.3} / \textbf{2.58} \\  %
        \bottomrule
	\end{tabular}
        \vspace{-0.3cm}\caption{\textbf{Results on NYUv2 dataset.} We report MSE and MAE metrics, the lower the better.}\vspace{-0.2cm} 
	\label{sota_comparison_mid_nyu_diml}
\end{table}

\subsection{Comparisons With State-of-the-Art Methods}
To assess the effectiveness of LFRD$^{2}$, we compare it with state-of-art methods, including traditional algorithms like CDNLM \cite{8293804}, JGDR \cite{rossi2020joint} and learning-based frameworks, namely ToFnet \cite{su2018deep}, ToF-KPN \cite{qiu2019deep},SHARPnet \cite{dong2020spatial}, PE-ToF \cite{chen2020very}, NAFNet~\cite{chen2022simple}, Restomer~\cite{zamir2022restormer} and UD-ToFnet \cite{qiao2022depth}. 

In Table \ref{sota_comparison}, we present quantitative results on the SUD-ToF and RUD-ToF datasets. Unsurprisingly, traditional methods perform worse than learning-based approaches.
Both LFRD$^{2}$ and UD-ToFnet~\cite{qiao2022depth} consistently outperform other methods, with LFRD$^{2}$ achieving better results than UD-ToFnet, except for the $\rho_{1.10}$ and $\rho_{1.02}$ scores on SUD-ToF and RUD-ToF, respectively.

We also report qualitative examples of depth maps restored by LFRD$^{2}$ compared to other learning-based methods. As shown by the error maps in Fig. \ref{fig:sota_comp1}, ToFnet \cite{su2018deep},  ToF-KPN \cite{qiu2019deep}, and PE-ToF \cite{chen2020very}  struggle to recover structural details, especially edges. Despite the improvement observed at edges for the result of SHARPnet \cite{dong2020spatial}, the irreversible loss of information inherent in the mapping from raw data to depth estimation methods poses challenges in achieving optimal depth estimation. Compared to other methods, our LFRD$^{2}$ demonstrates notable advantages in restoring depth quality and exhibits superior performance in edge-preserving, highlighting its effectiveness on the UD-ToF task.

\subsection{Beyond UD-ToF imaging.} Furthermore, we demonstrate how LFRD$^{2}$ is also effective for enhancing the quality of the depth maps obtained through classical ToF sensors -- e.g., not deployed under displays. Tab. \ref{sota_flat} shows results concerning ToF denoising on the FLAT dataset \cite{guo2018tackling}; it highlights that our framework significantly outperforms existing methods for this task. Additionally, Tab. \ref{sota_comparison_mid_nyu_diml} reports the results for the depth super-resolution task on the NYUv2 dataset \cite{Silberman:ECCV12}, performed according to three different upsampling factors -- $4\times$, $8\times$ and $16\times$. Once again, LFRD$^{2}$ achieves the best results with any upsampling factor, outperforming state-of-the-art DSR-EI \cite{qiao2023depth}. We refer the reader to the \textbf{supplementary material} for qualitative results.

\subsection{Ablation Analysis}

\begin{table}[t] \small   	
	\renewcommand\tabcolsep{4.5pt} 
	\centering
	\begin{tabular}{@{}ccccc@{}}
		\toprule
		\textbf{Method} & PE-ToF & NAFNet & Restomer & UD-ToFnet \\ \midrule
		Baseline & 21.2 / 48.7 & 20.4 / 39.8 & 18.9 / 31.8 & 17.3 / 31.1  \\
		LFRD$^2$ & \textbf{20.0} / \textbf{33.7} & \textbf{19.8} / \textbf{36.4} & \textbf{17.5} / \textbf{30.4} & \textbf{16.7} / \textbf{30.9}  \\
        \bottomrule
	\end{tabular}\vspace{-0.3cm}
        \caption{\textbf{Ablations on different baselines.} Our method adopts PE-ToF, NAFNet, Restomer, and UD-ToFnet as DISB to validate its effectiveness, with MAE / RMSE reported in the table.}\label{tab:baselines}
\end{table}

We now study in deeper detail the impact of the different modules composing our framework. For further details, see the \textbf{supplementary material}.

\textbf{Ablations on different baselines.}
In Table~\ref{tab:baselines}, we show on the RUD-ToF dataset how different existing models can be used as internal state builder, and how any of them get improved by our approach.

\textbf{Comparison with RNNs.} 
In Table~\ref{framework_ablation}, we analyze the results by LFRD$^2$ and the use of three Recurrent Neural Networks (RNNs), i.e., dilated convolution~\cite{Yu2015MultiScaleCA}, NLSPN~\cite{park2020non}, GRUs~\cite{cho2014learning} and LSTM~\cite{hochreiter1997long}, all of which were integrated with the baseline model for iterative optimization of depth.
"Dilated" refers to using dilated convolution with a fixed position rather than a flexible one to execute the diffusion process. 
NLSPN can be viewed as a diffusion process that employs deformable discrete convolution, exhibiting negligible performance gains compared to the baseline.
Gated recurrent units (GRUs) and Long short-term memory (LSTM) are RNN variants, with the latter being used in particular to capture long-term dependency; both attain improvements in terms of MAE, with negligible improvements -- or even drops -- in RMSE. We ascribe this to the higher dependency of LSTM on large amounts of training data.
Finally, although our goal is to privilege interpretability rather than outperform any alternative methods, we can appreciate that LRFD$^2$ consistently surpasses its counterparts, achieving state-of-the-art accuracy. 

Furthermore, Table 5 includes ablations without Continuous Convolutions (CC) or Fractional Calculus (FC), where “w/o FC” corresponds to the integer-order variant. Results show that both components are essential for optimal UD-ToF performance. We also compare with NFC~\cite{nsampi2023neural}, a continuous convolution baseline with similar accuracy but significantly higher FLOPs and runtime, highlighting the efficiency of our design.

\begin{table}[t] \small
	\centering
	\renewcommand\tabcolsep{3.4pt} 
	\begin{tabular}{@{}lccccc@{}} 
		\toprule
		\textbf{Config.} & \textbf{Params\,/\,M} & \textbf{Flops\,/\,G} & \textbf{Speed\,/\,ms} & \textbf{MAE} &  \textbf{RMSE}     \\
		\midrule
		Baseline & 2.17 &  8.65  & 15.20 & 17.29 & 31.11   \\
            Dilated  & $\approx 0$  & $\approx 0$  &  17.54 & 17.25  & 31.16   \\
            NLSPN   & +0.01 &  +3.23  & 17.70 & 17.23  & 31.14 \\ 
            GRU   & +0.18 &  +7.62  & 19.89 &  17.02  & 31.09  \\   
            LSTM   & +0.24 &  +10.4  & 22.15  &  16.96  & 31.22  \\  \midrule
            Ours    & +0.18 & +7.69 &  22.75 & \textbf{16.73} & \textbf{30.94} \\            
            \textit{w/o} FC  & +0.01 &  +0.41  & 20.67 & 17.00  & 30.99  \\ 
            \textit{w/o} CC  & +0.17 & +7.28 & 22.11 & 16.88 & 31.03   \\  
            NFC    & +0.13 & +20.5 & 28.42 & 16.97 & 31.00   \\
		\bottomrule
	\end{tabular}\vspace{-0.3cm}
        \caption{
        \textbf{Ablation study -- LFRD$^2$ main components.} We compare LFRD$^2$ with RNN variants (top) and evaluate the impact of key modules -- Fractional Calculus (FC), Continuous Convolution (CC), and NFC~\cite{nsampi2023neural} -- in the iterative process (bottom).
        }\vspace{-0.3cm}
	\label{framework_ablation}
\end{table}

\textbf{Ablations on Continuous Convolution.} 
This section presents how inputs of continuous convolution affect the performance of LFRD$^2$. We compare the outcomes of employing different concatenations of depth, intermediate features of depth (Simplified as features), amplitude, and offsets during the iteration. As shown in Table \ref{ablation_cc}, 
we notice that concatenating amplitude and features results in optimal performance, whereas concatenating four elements fails to yield better results. 
Upon analysis, we ascribe this to the fact that during the iterative process, the depth tends to steer the attention of the network toward previous depth states, compromising final accuracy. 
Therefore, we select amplitude, features, and offsets as inputs to be forwarded to our continuous convolution modules.

\begin{table}[t]	    
	\centering
	\renewcommand\tabcolsep{5pt} 
	\begin{tabular}{@{}cccccc@{}}
		\toprule
		   & \textbf{Depth} & \textbf{Features} & \textbf{Amplitude} & \textbf{MAE} & \textbf{RMSE}  \\
		\midrule	
            \textbf{\uppercase\expandafter{\romannumeral1}}  &   \XSolidBrush & \XSolidBrush & \Checkmark & 17.19 & 31.20   \\            
            \textbf{\uppercase\expandafter{\romannumeral2}}  &   \XSolidBrush  & \Checkmark & \XSolidBrush & 17.16 & 31.22 \\
            \textbf{\uppercase\expandafter{\romannumeral3}}  &   \Checkmark & \Checkmark &  \Checkmark  & 16.91 & 30.98 \\
		  \textbf{\uppercase\expandafter{\romannumeral4}}  &   \XSolidBrush &   \Checkmark & \Checkmark   & \textbf{16.73} & \textbf{30.94} \\		  
		\bottomrule
	\end{tabular}\vspace{-0.3cm}
 \caption{\textbf{Ablation study -- Continuous Convolution.} Comparison among four variants using different inputs.}\vspace{-0.3cm}
	\label{ablation_cc}
\end{table}

\textbf{Effect of Fractional Order Selection.} Following FF~\cite{zamora2021convolutional}, we compare our variable fractional order selection with different fixed orders ranging from 0 to 1. Tab.~\ref{variable_order} shows that as the fractional order varies, the model performance undergoes notable changes in accuracy. 
Although our method achieves only marginal improvement over fixed orders of $0.1$ and $0.2$ in terms of $\rho_{1.02}$, it significantly outperforms them in the MAE metric. Overall, our variable order yields the optimal results.

\begin{table}[t]	    
	\centering
	\renewcommand\tabcolsep{4pt} 
	\begin{tabular}{@{}lcccccc@{}}
		\toprule
		  \textbf{Order} & 0.1 & 0.3 & 0.5 & 0.7 & 0.9 & Ours  \\
		\midrule	
             {MAE/mm} &  18.12  & 18.38 & 18.86 &  19.01 & 18.29 & \textbf{17.62} \\ 
            {$\sigma_{1.02}$ /\%} &  66.79 & 66.80 & 66.16 & 65.73 & 66.04 & \textbf{67.43} \\  
		\bottomrule
	\end{tabular}\vspace{-0.3cm}
 \caption{\textbf{Ablation study -- fractional order.} Comparison between models with fixed fractional orders and ours.}\vspace{-0.3cm}
	\label{variable_order}
\end{table}

\begin{figure}[t] 
	\centering
	\renewcommand\tabcolsep{1.5pt} 
	\begin{tabular}{ccccc}
	\vspace{-0.1cm}
        \includegraphics[width=0.8in]{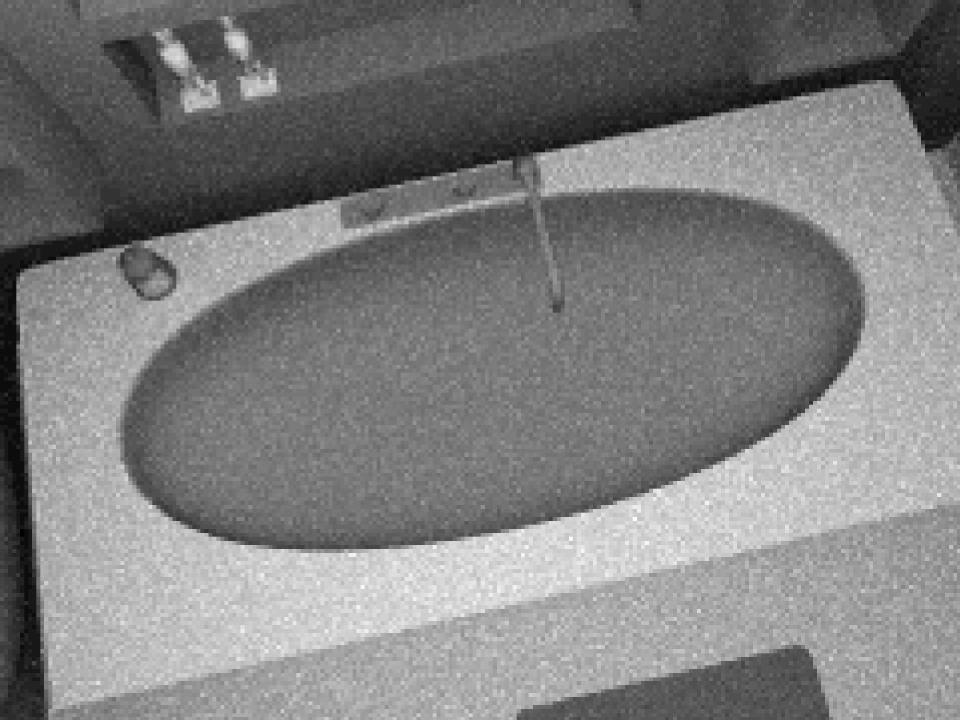}
	\hspace{-1.8mm} & \includegraphics[width=0.8in]{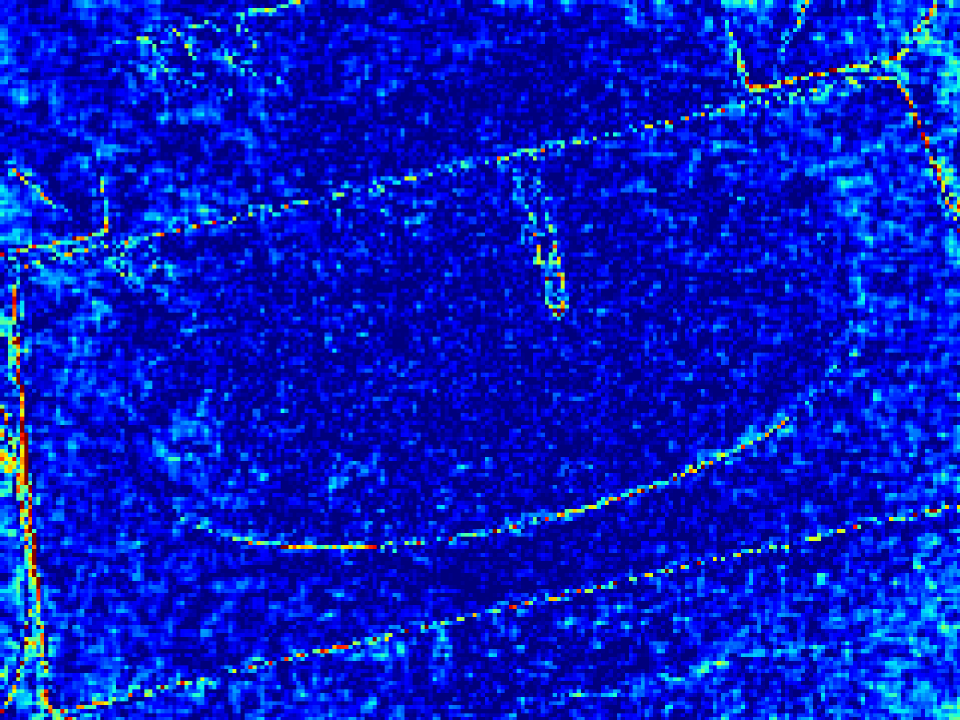}
	\hspace{-1.8mm} & \includegraphics[width=0.8in]{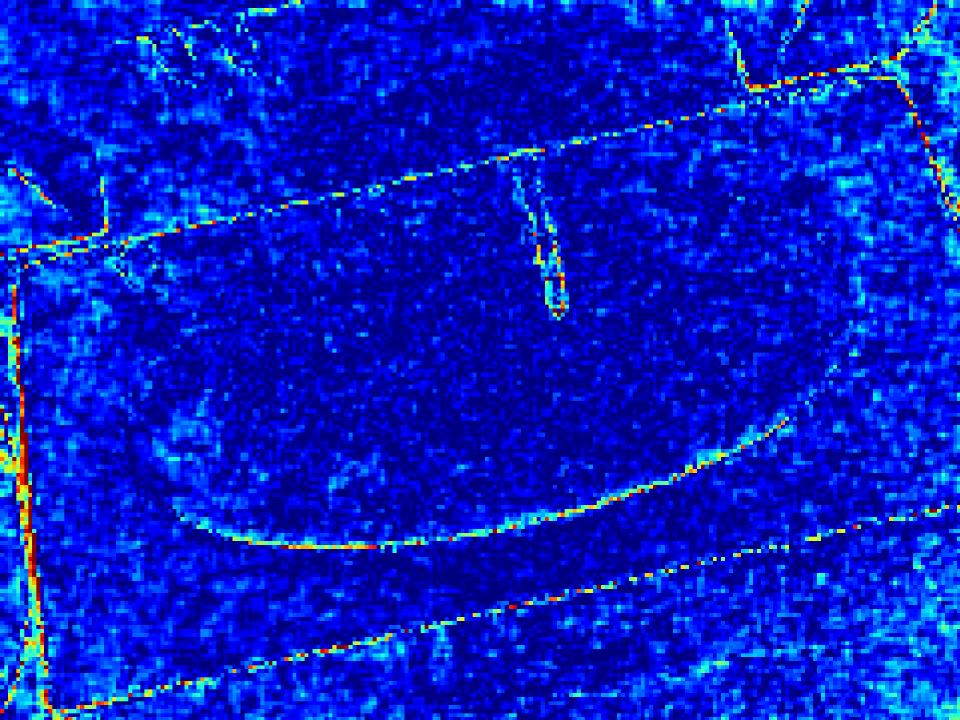}
        \hspace{-1.8mm} &\includegraphics[width=0.8in]{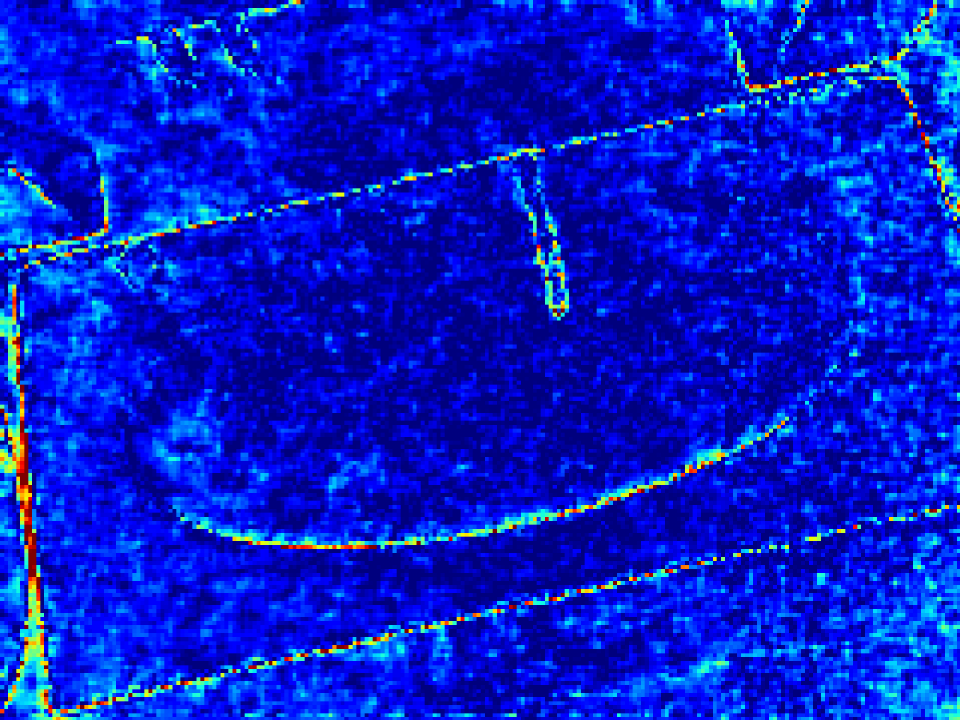}
	 \\
	\small  IR  & \small  $E_0$  & \small  $E_1$ & \small  $E_2$ 
        \\
         \includegraphics[width=0.8in]{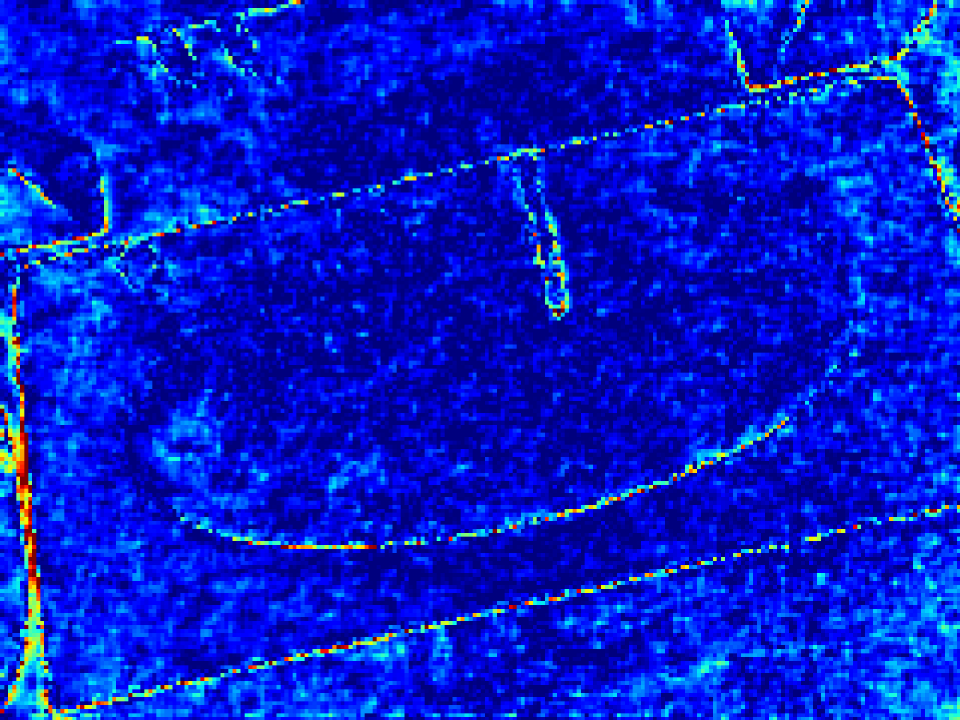}
         \hspace{-1.8mm} & \includegraphics[width=0.8in]{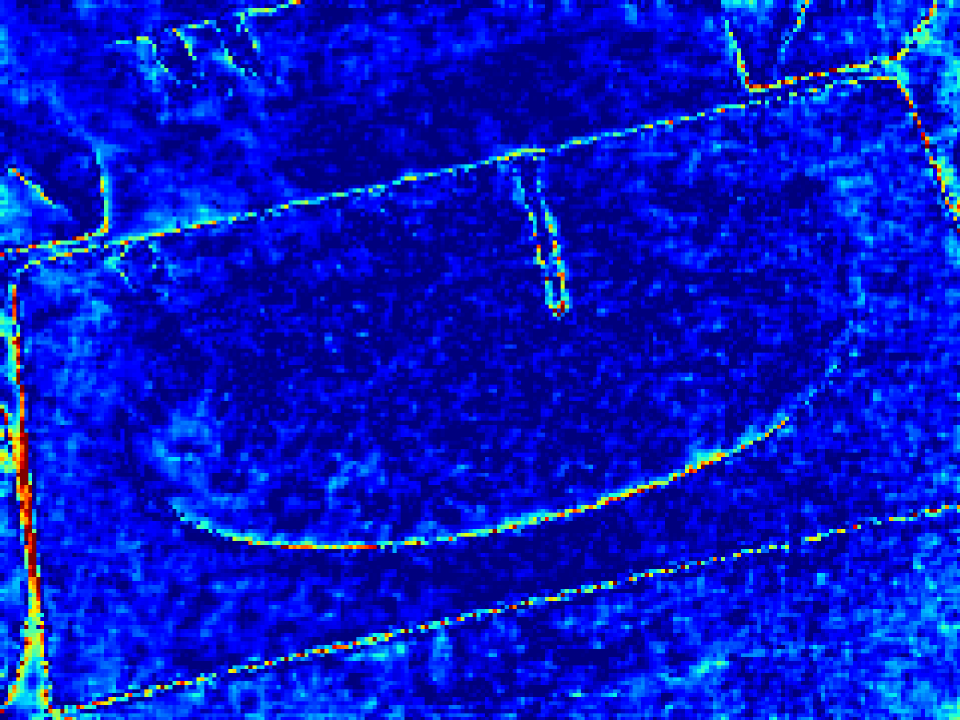}
	\hspace{-1.8mm} & \includegraphics[width=0.8in]{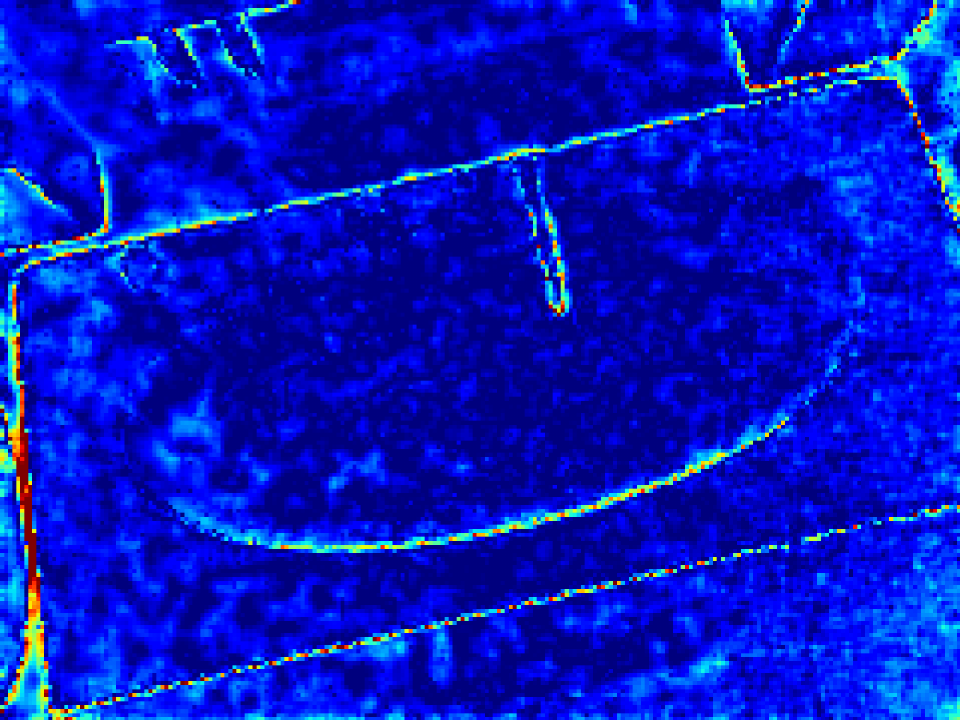}
	\hspace{-1.8mm} & \includegraphics[width=0.8in]{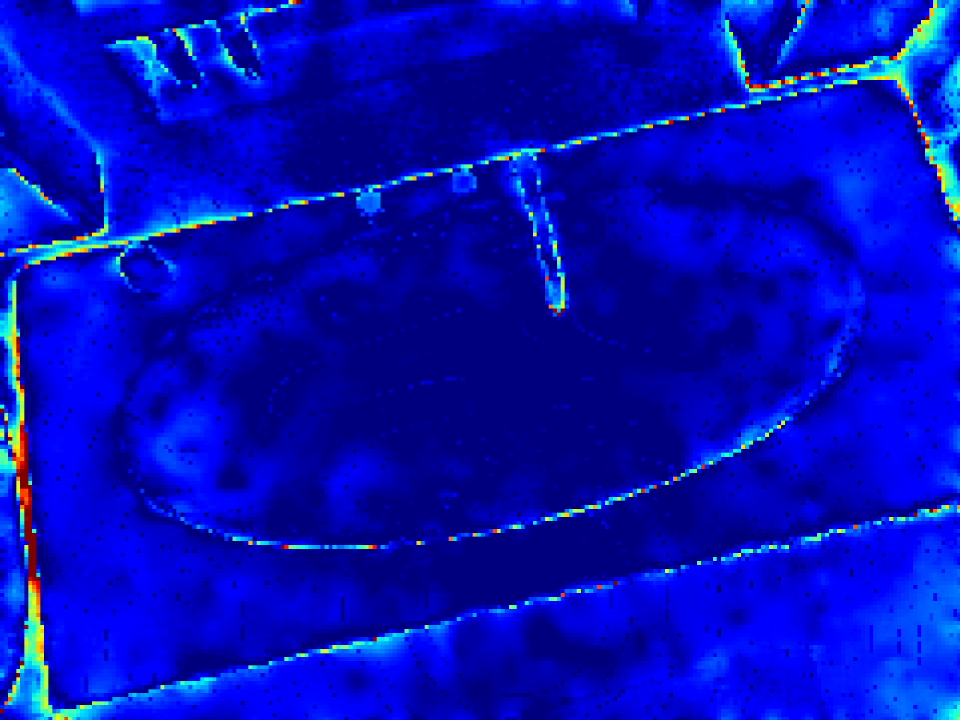}
    \\
    \small  $E_3$ & \small  $E_4$ & \small  $E_5$ & \small  $E_6$ 
         \\ \vspace{-0.1cm}
    
	\end{tabular}
    \vspace{-0.7cm}
	\caption{\textbf{Qualitative results of different iterations.} IR represents the IR image, and $E_i$ denotes the error map after $i$-th iteration.} 
	\label{fig:iteration}
\end{figure}

\textbf{Qualitative Results at Different Iterations}
To evaluate the effectiveness of our iterative process, we present the error maps at different iterations. In Fig. \ref{fig:iteration}, as the iteration number increases, depth consistently and progressively approximates the ground truth depth, indicating a steady improvement in accuracy toward the desired outcome.

\subsection{Limitations}
Despite the higher interpretability, it is crucial to carefully set the training strategy to avoid instabilities -- i.e., NaN values. 
In the future, we aim to develop a more robust method and validate its effectiveness across various tasks for depth restoration.
Besides, we plan to devise a more efficient hybrid architecture based on the implicit numerical scheme.

\section{Conclusion}
In this paper, we proposed a physical model-driven deep framework for UD-ToF depth restoration, which integrates the underlying physical knowledge into a convolutional neural network, thereby iteratively facilitating the learning of spatio-temporal dynamics from the depth information. 
This approach encodes the time-fractional reaction-diffusion equation into the designed neural module, endowing the diffusion process with long-term memory properties.
To further enhance depth quality, an efficient non-local continuous convolution operator is introduced. 
This operator enables the framework to achieve continuous convolution in a discrete form by predicting coefficients based on linear approximation and repeated differentiation.
The experimental results demonstrate that our framework not only leverages the powerful representation learning capabilities of neural networks but also respects the underlying physics, resulting in more accurate and robust UD-ToF imaging.

{
    \small
    \bibliographystyle{ieeenat_fullname}
    \bibliography{main}
}

\end{document}